%% file: main_arxiv.tex
\documentclass[10pt,twocolumn,letterpaper]{article}

\usepackage[pagenumbers]{cvpr} 

\usepackage{graphicx}
\usepackage{amsmath}
\usepackage{amssymb}
\usepackage{booktabs}

\usepackage{algorithm}
\usepackage{algpseudocode}
\usepackage{multirow}

\usepackage{caption}
\usepackage{subcaption}

\usepackage[pagebackref,breaklinks,colorlinks]{hyperref}
\usepackage[accsupp]{axessibility}  

\usepackage[capitalize]{cleveref}
\crefname{section}{Sec.}{Secs.}
\Crefname{section}{Section}{Sections}
\Crefname{table}{Table}{Tables}
\crefname{table}{Tab.}{Tabs.}

\newcommand{\moniker}{SinGRAF}

\begin{document}

\input{camera/sec0.tex}

\input{camera/sec1.tex}

\input{camera/sec2.tex}
\input{camera/sec3.tex}
\input{camera/sec5.tex}

\input{camera/sec6.tex}

\input{camera/ack.tex}

{\small
\bibliographystyle{ieee_fullname}
\bibliography{egbib}
}

\clearpage
\appendix
\section*{\Large{Supplementary}}
\input{camera/supp1.tex}

\input{camera/supp3.tex}

\input{camera/supp2.tex}

\end{document}

%% file: camera/sec0.tex
\title{SinGRAF: Learning a 3D Generative Radiance Field for a Single Scene}

\author{Minjung Son\thanks{Equal contribution.\newline Project page: \href{http://www.computationalimaging.org/publications/singraf/}{computationalimaging.org/publications/singraf/}}\, $^{1,2}$ \qquad Jeong Joon Park\footnotemark[1]\, $^2$ \qquad Leonidas Guibas$^2$ \qquad Gordon Wetzstein$^2$ \\
$^1$Samsung Advanced Institute of Technology (SAIT) \hspace{1em} $^2$Stanford University\\
{\tt\small minjungs.son@samsung.com, \{jjpark3d,gordon.wetzstein,guibas\}@stanford.edu}
}
\maketitle

\begin{abstract}
Generative models have shown great promise in synthesizing photorealistic 3D objects, but they require large amounts of training data. We introduce SinGRAF, a 3D-aware generative model that is trained with a few input images of a single scene. Once trained, SinGRAF generates different realizations of this 3D scene that preserve the appearance of the input while varying scene layout. For this purpose, we build on recent progress in 3D GAN architectures and introduce a novel progressive-scale patch discrimination approach during training. With several experiments, we demonstrate that the results produced by SinGRAF outperform the closest related works in both quality and diversity by a large margin.  
\end{abstract}

%% file: camera/sec1.tex
\section{Introduction}
\label{sec:intro}

Creating a new 3D asset is a laborious task, which often requires manual design of triangle meshes, texture maps, and object placements. 
As such, numerous methods were proposed to automatically create diverse and realistic variations of existing 3D assets. For example, procedural modeling techniques \cite{ebert2003texturing, mvech1996visual} produce variations in 3D assets given predefined rules and grammars, and example-based modeling methods \cite{funkhouser2004modeling, kalogerakis2012probabilistic} combine different 3D components to generate new ones.
 
With our work, we propose a different, generative strategy that is able to create realistic variations of a single 3D scene from a small number of photographs. Unlike existing 3D generative models, which typically require 3D assets as input~\cite{funkhouser2004modeling, wu2022learning}, our approach only takes a set of unposed images as input and outputs a generative model of a single 3D scene, represented as a neural radiance field~\cite{mildenhall2021nerf}.

\begin{figure}[t!]
    \centering
    \includegraphics[width=.47\textwidth]{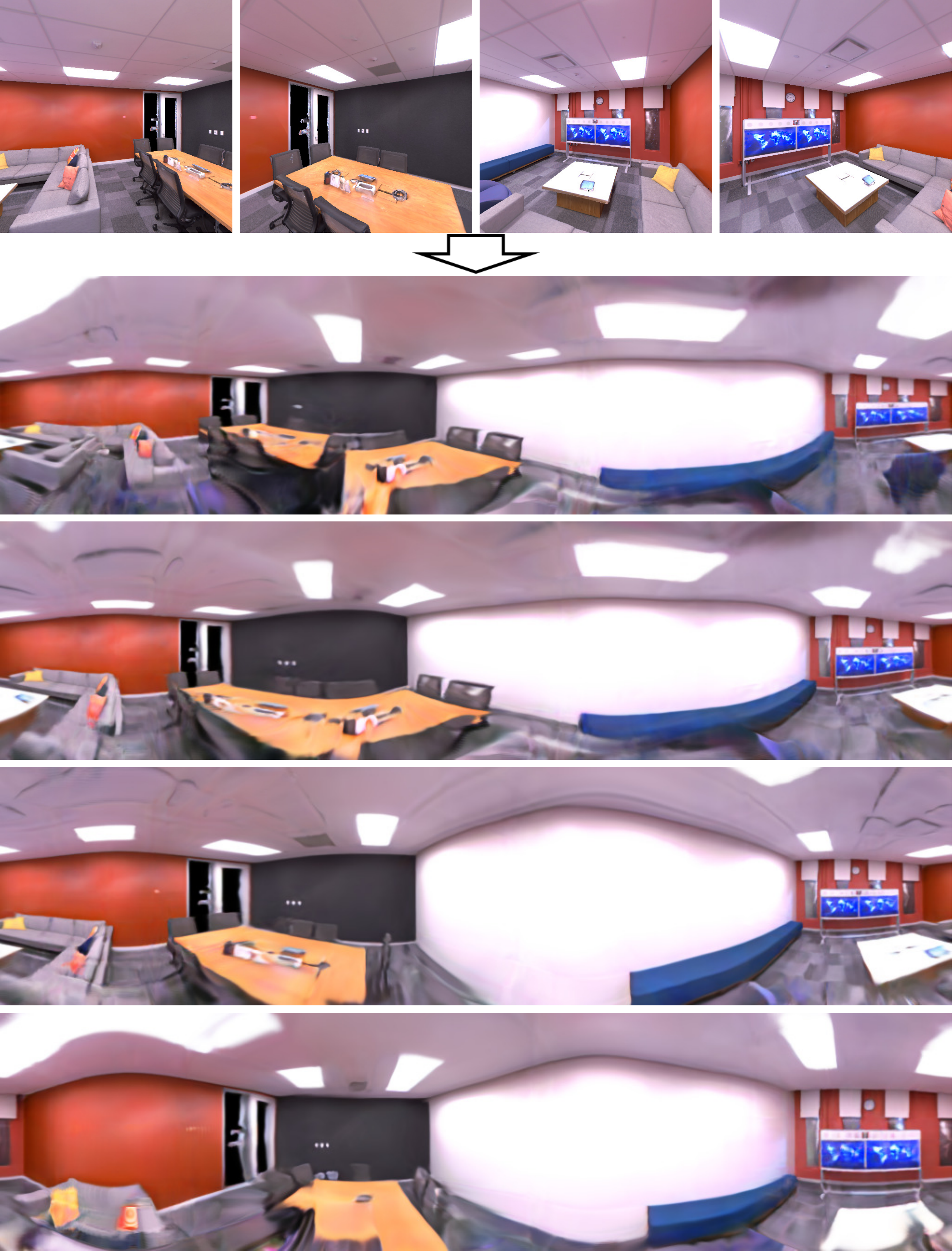}
    \caption{\moniker{} generates different plausible realizations of a single 3D scene from a few unposed input images of that scene. In this example, i.e., the ``office\_3'' scene, we use 100 input images, four of which are shown in the top row. Next, we visualize four realizations of the 3D scene as panoramas, rendered using the generated neural radiance fields. Note the variations in scene layout, including chairs, tables, lamps, and other parts, while staying faithful to the structure and style of the input images.}
    \label{fig:teaser}
\end{figure}

Our method, dubbed \moniker, builds on recent progress in unconditional 3D-aware GANs~\cite{chan2021pi, schwarz2020graf} that train generative radiance fields from a set of single-view images. However, directly applying these 3D GANs to our problem is challenging, because they typically require a large training set of diverse images and often limit their optimal operating ranges to objects, rather than entire scenes. \moniker{} makes a first attempt to train a 3D generative radiance field for individual indoor 3D scenes, creating realistic 3D variations in scene layout from unposed 2D images.

Intuitively, our method is supervised to capture the internal statistics of image patches at various scales and generate 3D scenes whose patch-based projections follow the input image statistics.
At the core of our method lies continuous-scale patch-based adversarial \cite{goodfellow2020generative} training. Our radiance fields are represented as triplane feature maps~\cite{Chan2022, Shue2023triplanediffusion} produced by a StyleGAN2~\cite{karras2020analyzing} generator. We  volume-render our generated scenes from randomly sampled cameras with {\em varying fields of view}, to simulate the appearance of image patches at various scales. A scale-aware discriminator is then used to compute an adversarial loss to the real and generated 2D patches to enforce realistic patch distributions across all sampled views. Notably, our design of continuous-scale patch-based generator and discriminator allows patch-level adversarial training without expensive hierarchical training \cite{Shaham_2019_ICCV,wu2022learning, Xu_2021_CVPR}. During the training, we find applying perspective augmentations to the image patches and optimizing the camera sampling distribution to be important for high-quality scene generation.

The resulting system is able to create plausible 3D variations of a given scene trained only from a set of unposed 2D images of that scene. We demonstrate our method on two challenging indoor datasets of Replica~\cite{straub2019replica} and Matterport3D~\cite{chang2017matterport3d} as well as a captured outdoor scene. We evaluate \moniker{} against the state-of-the-art 3D scene generation methods, demonstrating its unique ability to induce realistic and diverse 3D generations.

%% file: camera/sec2.tex
\section{Related Work}
\label{sec:related}

\paragraph{Synthesis from 3D Supervision.} A large body of prior work aims at creating variations of scenes or objects. Procedural modeling approaches~\cite{ebert2003texturing, parish2001procedural, mvech1996visual, musgrave1989synthesis} are widely used for auto-generating repetitive scenes such as terrains, buildings, or plants. These methods typically require manually designing the rules and grammars to procedurally add new 3D elements. Example-based methods \cite{kalogerakis2012probabilistic, funkhouser2004modeling, xu2012fit} aim at extracting patterns from 3D asset examples to synthesize new models.  This line of data-driven approaches learns how to mix and match different components to create a plausible 3D asset. Similarly, scene synthesis techniques \cite{paschalidou2021atiss, fisher2012example, wang2018deep} learn a distribution of plausible object arrangements from professionally designed scene datasets. All of these methods require datasets of 3D assets, part segmentation, or object arrangement designs, which are expensive to collect.

\paragraph{3D-aware GANs.} 
Leveraging the recent developments of neural implicit representations \cite{park2019deepsdf,mescheder2019occupancy,chen2019learning,sitzmann2019siren} and radiance fields \cite{sitzmann2019srns,mildenhall2021nerf, barron2021mip, lindell2022bacon,muller2022instant}, 
3D GANs \cite{chan2021pi,niemeyer2021giraffe,Chan2022,gu2021stylenerf,or2022stylesdf, xu20223d, zhou2021cips,DeVries_2021_ICCV,bautista2022gaudi,shi20223daware,skorokhodov2022epigraf,xue2022giraffehd,zhang2022mvcgan,deng2022gram,xiang2022gramhd,Tewari_2022_CVPR,bahmani20223d} train generative 3D radiance fields from a set of single-view images. These methods render the sampled scenes from various viewpoints via volume rendering and supervise adversarially. Many of these approaches apply their discriminators on full-resolution images during training, but some also employ patch-based discriminators~\cite{schwarz2020graf,meng2021gnerf,skorokhodov2022epigraf}. The resulting 3D GANs can create diverse 3D radiance fields that enable view-consistent NVS. Existing 3D GANs, however, rely on a large amount of training data, while our model only uses a few images of a single 3D scene.

\paragraph{Few-Shot Generative Models.} 
Recently, researchers have started applying generative modeling techniques to few-shot settings, where only a few or single examples are given. 
In the 2D image domain, SinGAN and its extensions \cite{Xu_2021_CVPR,Shaham_2019_ICCV,hinz2021improved} explored the idea of training a CNN-based hierarchical generator on a single image, supervised using patch discrimination at multiple scales. The strategy of learning the internal patch distribution of a single example to train a generative model has been widely adopted for various tasks, including the synthesis of videos \cite{haim2022diverse}, motion sequences \cite{li2022ganimator}, 3D textures \cite{henzler2020learning, portenier2020gramgan}, or 3D shapes \cite{hertz2020deep,wu2022learning}. 
However, none of these works applied 3D generative models from single-scene images. Concurrently to our work, \cite{wang2022singrav} trains generative radiance fields from single-scene images but focuses mainly on stochastic and repetitive synthetic scenes, rather than structured, human-made scenes.

%% file: camera/sec3.tex
\begin{figure}[t!]
    \centering
    \includegraphics[width=\columnwidth]{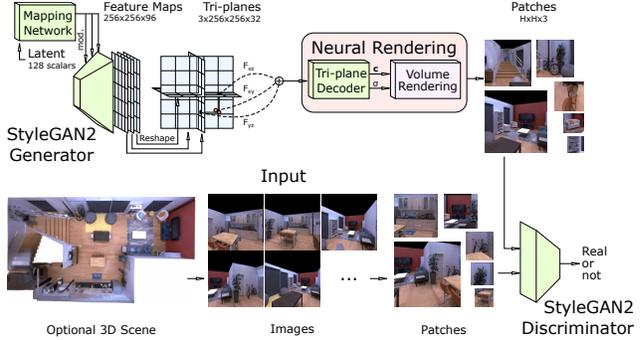}
    \caption{\moniker{} pipeline. The framework takes as input a few images of a single scene, for example rendered from a 3D scan or photographed (bottom). The 3D-aware generator (top left) is then trained to generate 2D feature planes that are arranged in a triplane configuration and rendered into patches of varying scale (top right). These rendered patches along with patches cropped from the input images are then compared by a discriminator. Once trained, the \moniker{} generator synthesizes different realizations of the 3D scene that resemble the appearance of the training images while varying the layout.}
    \label{fig:pipeline}
\end{figure}

\section{Single Scene 3D GAN}
\label{sec:approach}

Our system takes as input an unposed set of images  taken from a single scene and outputs a 3D generative model $\mathbf{G}$ that can generate diverse 3D radiance fields. We assume the intrinsic parameters of the camera are available. We do not assume our target scene to be static, i.e., we allow temporal changes in the scene. In the experiment section, we show our method's behavior for dynamic scenes.
We illustrate our image generation and discrimination processes in Fig.~\ref{fig:pipeline}. Source code and pre-trained models will be made available.

\subsection{Rendering Model}
Our generative model $\mathbf{G}(\mathbf{\Gamma}, \mathbf{z})$ takes a set of query rays $\mathbf{\Gamma}$ and a noise vector $\mathbf{z}$ and outputs RGB predictions for the rays. Below we briefly discuss the rendering process of $\mathbf{G}.$

\paragraph{Generator Architecture}
$\mathbf{G}$ represents continuous radiance fields using tri-planes, following~\cite{Chan2022}. We adopt a StyleGAN2-based generator backbone ~\cite{karras2020analyzing}, which 
consists of a mapping network that takes as input a noise vector $\mathbf{z} \sim \mathbb{R}^{128}$ and transforms it into a latent code vector $\mathbf{w} \sim \mathbb{R}^{128}$. Next, a synthesis network transforms $\mathbf{w}$ into a 2D feature image $\mathbf{F} \in \mathbb{R}^{N \times N \times 3C}$ with a total of $3C$ feature channels. Following~\cite{Chan2022}, we split these feature channels into three axis-aligned feature planes $\mathbf{F}_{xy},\mathbf{F}_{xz},\mathbf{F}_{yz} \in \mathbb{R}^{N \times N \times C}$.   

A color $\mathbf{c}$ and density $\sigma$ value can now be queried at an arbitrary 3D coordinate $\mathbf{x}$ by aggregating the triplane features and processing them by a small multilayer perceptron--style decoder, $\textrm{\sc{mlp}}: \mathbb{R}^{3C} \rightarrow \mathbb{R}^4$, as
\begin{equation}
    \left( \mathbf{c} \left( \mathbf{x} \right), \sigma \left( \mathbf{x} \right) \right)  = \textrm{\sc{mlp}} \left( \mathbf{F}_{xy} \left(\mathbf{x}\right) \! + \!  \mathbf{F}_{xz} \left(\mathbf{x}\right) \! + \! \mathbf{F}_{yz} \left(\mathbf{x}\right) \right).
\end{equation}
Note that we do not model view-dependent effects.

\paragraph{Neural Rendering.}

Using volume rendering~\cite{Max:1995,mildenhall2021nerf}, we project the 3D neural field into 2D images. For this purpose, the aggregated color $\mathbf{C}(\mathbf{r})$ of a ray $\mathbf{r}\in \mathbf{\Gamma}$ is computed by integrating the field as
\begin{align}
\mathbf{C} (\mathbf{r}) & = \int_{t_{n}}^{t_{f}} T \left( t \right) \sigma \left( \mathbf{r} (t) \right) \mathbf{c} \left( \mathbf{r}(t) \right) \textrm{d} t, \\
T \left( t \right) & = \textrm{exp} \left( - \int_{t_{n}}^{t} \sigma \left( \mathbf{r} \left( s \right) \right) \textrm{d} s \right),
\label{eq:volumerendering}
\end{align}
where $t_n$ and $t_f$ indicate near and far bounds along the ray $\mathbf{r} (t) = \mathbf{o} + t \mathbf{d}$ pointing from its origin $\mathbf{o}$ into direction $\mathbf{d}$. The continuous volume rendering equation (\cref{eq:volumerendering}) is typically computed using the quadrature rule~\cite{Max:1995}.

Our volume rendering step directly outputs RGB color images or patches; we do not apply a superresolution module on the rendered values. Moreover, we use 96 samples per ray and do not use hierarchical ray sampling \cite{mildenhall2021nerf}.

\subsection{Training Process} \label{sec: training}
\paragraph{Progressive Patch Scaling.}
Given a set of input images, most existing 3D GANs generate and discriminate images at full resolution. However, as shown in the ablation study (Sec.~\ref{sec:analysis}), adversarial training at full scale leads to a collapse of the learned distribution to a single mode, likely because the joint information of the images uniquely determines a single 3D structure. Therefore, we turn to learn the internal patch distribution of the images, inspired by some existing works \cite{Shaham_2019_ICCV, wu2022learning, Xu_2021_CVPR}. 

Directly extending hierarchical patch-based GANs \cite{Shaham_2019_ICCV,wu2022learning, Xu_2021_CVPR}, however, would require expensive training of a pyramid of generators with progressively-growing feature plane resolutions. To address the issue, we notice that our tri-plane features define a continuous radiance field and thus are able to render patches at an arbitrary resolution and scale. Therefore, we use a single generator network and continuously control the scale of the patches during training, forgoing the progressive training of multiple generator networks \cite{Shaham_2019_ICCV,Xu_2021_CVPR,wu2022learning}.

\begin{figure}[t!]
    \centering
    \includegraphics[width=.47\textwidth]{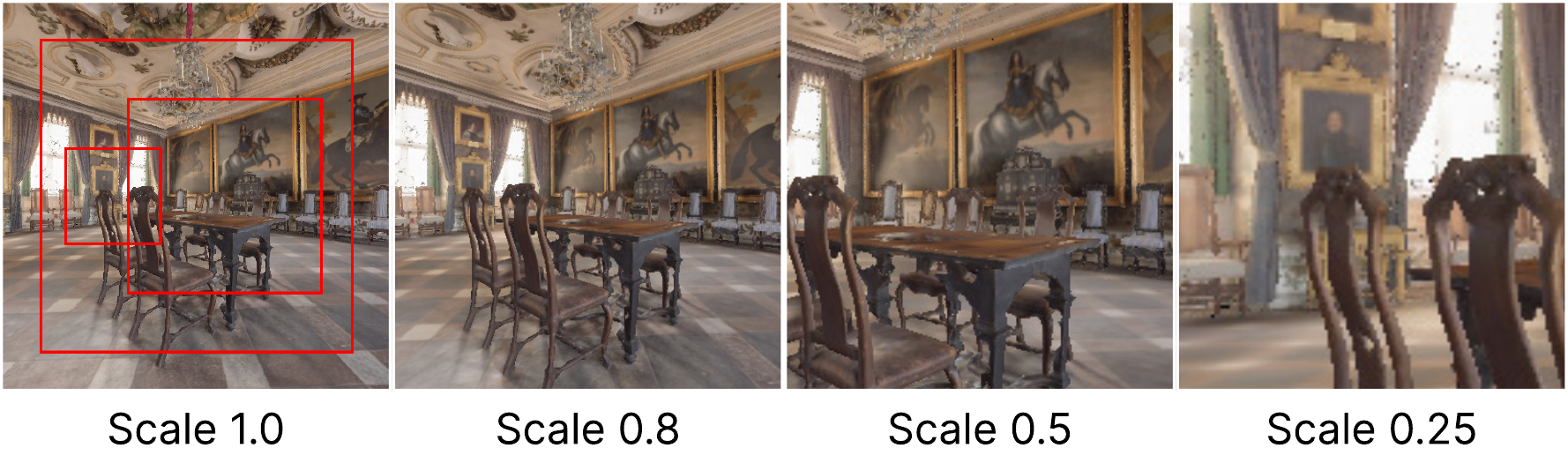}\\
    \caption{Visualizations of patches at different scales.}
    \label{fig:patch_scale}
\end{figure}

Given input images $\{I_1, ... ,I_N | I_i \in \mathbb{R}^{H'\times H'\times 3}\}$ of resolution $H' \times H'$ and field of view (FOV) $\theta$, we consider the image plane $\mathcal{P}$ of a virtual camera with the same FOV. 
During training, we volume-render patches with {\em fixed} resolution $H \times H$ but with varying scale $s$. Here, the scale $s\in [0,1]$ indicates the spatial extent of a  patch on the image plane of $\mathcal{P}.$ When $s=1.0$ the patch covers the entire image plane with $H\times H$ pixels, which is used for full-resolution training of existing 3D GANs \cite{chan2021pi,schwarz2020graf}. With smaller $s,$ the patch will cover a smaller window in the image plane with the same resolution, thus containing more details. The location of the patch window on $\mathcal{P}$ is sampled randomly. We render a sampled patch $\rho$ with its associated rays $\mathbf{\Gamma}_\rho$: $\mathbf{G}(\mathbf{\Gamma}_\rho,\mathbf{z}).$ Similarly, a ground truth patch can be sampled by cropping $I_i$'s with a given patch scale $s$ (see Fig.~\ref{fig:patch_scale}).

The scale value $s$ is sampled randomly for each patch from a uniform distribution $s\sim \texttt{U}(s_{\min}(t), s_{\max}(t)),$ where $t$ is the current training epoch. We schedule the scale distribution so that in early epochs, we have larger patches to provide scene structural information and gradually decrease the scales towards the end of the training to induce better quality and diversity. Refer to supplementary for details.

Our discriminator network $\mathbf{D}$, whose architecture closely follows that of \cite{karras2020analyzing}, takes as input patches and outputs a scalar, indicating the realism of the patches. Because our patches have varying scales, we additionally condition $\mathbf{D}$ with the scale value $s$ of each patch $\rho$:
\begin{equation}
    \mathbf{D}: \mathbf{G}(\mathbf{\Gamma}_\rho,\mathbf{z}) \times s \mapsto \mathbb{R}.
\end{equation}
Implementation-wise, we simply repeat the scale value to match the patch resolution.

\paragraph{Data augmentation.}

Our approach aims at generating a 3D scene from a limited set of 2D observations. As such, it is desirable to augment the available data during training. To this end, we apply data augmentation techniques for both image and camera pose data.

In addition to the usual image augmentation techniques used for 2D GANs, such as translation, cropping, and cutout, we suggest a perspective augmentation approach, which is appropriate for generating 3D scenes with patch discrimination. Based on the known camera pose, we re-project image patches so as to imitate camera rotations followed by perspective projections. Camera rotations without changing the position do not induce occlusions or parallax; thus, this approach is applicable to captured content. Moreover, because our model discriminates patches instead of full images, unknown regions after rotation and perspective projection can be cropped. In practice, we start our training without rotational augmentation but gradually introduce and increase this augmentation up to $15^\circ$ for later training epochs where the patch scale $s$ decreases. Additional details on data augmentation are included in the supplement.

\paragraph{Camera Pose Distribution.}
To render patches from radiance fields we need to sample a virtual camera to render from. We define the camera pose distribution non-parametrically using a set of 1,000 cameras: $\mathcal{T}=\{T_1,...,T_{1000}|T_i\in SE(3)\}$; each virtual camera $\tau$ is randomly chosen from $\mathcal{T}$ during training. 
We reject the sampled $\tau$ when the occupancy (opacity) value at the camera center is above some threshold.  The $T_i$'s are initially sampled from a zero-mean 2D Gaussian on a plane with shared heights with random rotation about the vertical axis. During training, we jitter the $T_i$'s on translation and rotation.

The randomly initialized camera distribution, however, may not be a good representation of the real camera distribution of the input images. While our method can generate reasonable 3D scenes when training with random distributions, we find it beneficial to optimize the camera distribution $\mathcal{T}$ for higher-quality outputs. Such optimization will transform the distribution to accommodate poses with various rotations. Therefore, we optimize the $T_i$'s via automatic differentiation using our adversarial loss in the early stage of the training. We eventually stop the optimization since smaller patches at the later stage contain weak pose information. In practice, we decompose the poses into more easily optimizable forms, which we discuss further in the supplementary.

\paragraph{Training Objective}
Besides the regular min-max adversarial loss \cite{goodfellow2020generative}, we find it useful to adopt two regularization losses: R1 gradient penalty \cite{mescheder2018training} and discriminator reconstruction loss \cite{liu2020towards} ($\mathcal{E}_{R}$). Combining the regularizations with the adversarial loss, we have the final objective as follows:
\begin{align}
\mathcal{E}(\mathbf{G},&\mathbf{D},\mathcal{T})=\mathbb{E}_{\mathbf{z},\tau \sim \mathcal{T},s,\rho}\left[ f(\mathbf{D}(\mathbf{G}(\mathbf{\Gamma}_\rho,\mathbf{z}))) \right]+\\
&\mathbb{E}_{I,s,\rho}\left[f(-\mathbf{D}(\rho,s)) +
\lambda_{1} |\nabla_\rho \mathbf{D}(\rho,s)|^2  \right]+\lambda_{2}\mathcal{E}_{R}(\mathbf{D}), \nonumber
\end{align}
where $f(a)=-\log(1+\exp(-a)),$ and $\lambda$'s are balancing parameters. Note that the sampled patch $\rho$ is a function of the sampled scale $s$. We minimize the objective using the ADAM optimizer with learning rate of 2e-3. 

%% file: camera/sec5.tex
\section{Experiments}
\label{sec:results}

We conduct a number of experiments to test SinGRAF's capabilities to (i) create realistic 3D scene variations across a diverse set of challenging indoor scenes, (ii) reliably produce view-consistent 3D representations, and (iii) handle scene dynamics. At the end of the section, we show ablation studies and justify our design decisions. We refer  to supplementary for additional empirical analysis and results.

\subsection{Datasets}
We test and compare our approach on five scenes from the Replica dataset and one scene from the Matterport3D dataset, featuring realistically scanned indoor scenes. Within the Replica dataset, we choose to cover a good range of different scene types, including offices, apartments, and hotel rooms. The Replica dataset, however, only contains scanned indoor scenes that are typical of their own categories. To stress-test our algorithm, we add a large `castle' ballroom dataset from the Matterport3D dataset that has a very different appearance from `regular' indoor scenes. For each of the above 6 scenes, we render 100 views by sampling camera locations from a zero-mean Gaussian distribution with random rotations about the height axis. We reject the cameras that collide with occupied volumes. Lastly, we showcase our method on a captured image dataset where we use our own photographs of an outdoor scene. We capture the images using a hand-held consumer-grade smartphone camera and pass them to our algorithm without intrinsic or extrinsic calibrations.

\begin{figure*}
     \centering
     \begin{subfigure}[b]{0.49\textwidth}
        \includegraphics[width=\linewidth]{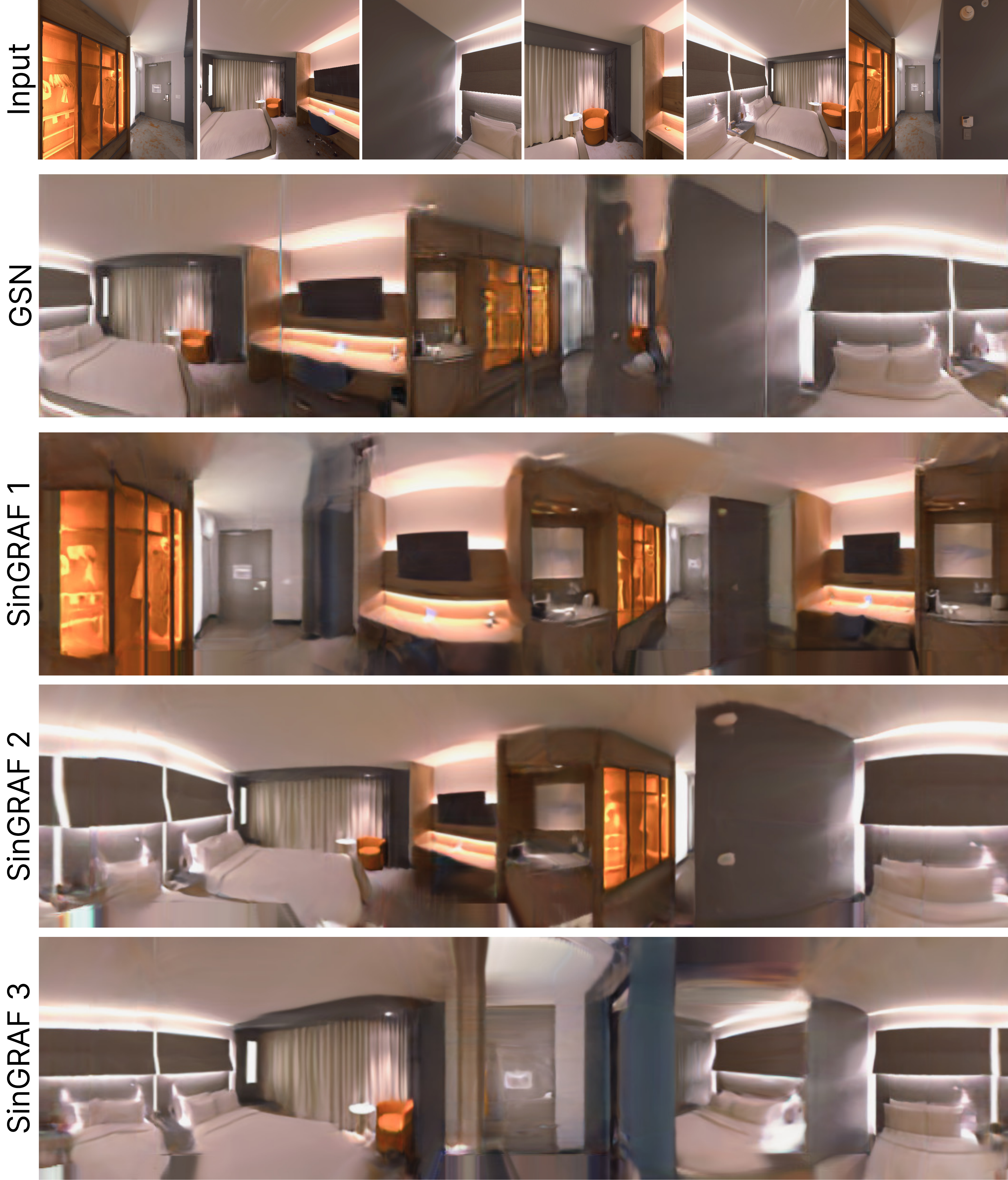}
        \caption{``hotel\_0'' scene from the Replica dataset. 
        }
        \label{fig:results_hotel_0}
    \end{subfigure}
     \hfill
     \begin{subfigure}[b]{0.49\textwidth}
        \includegraphics[width=\linewidth]{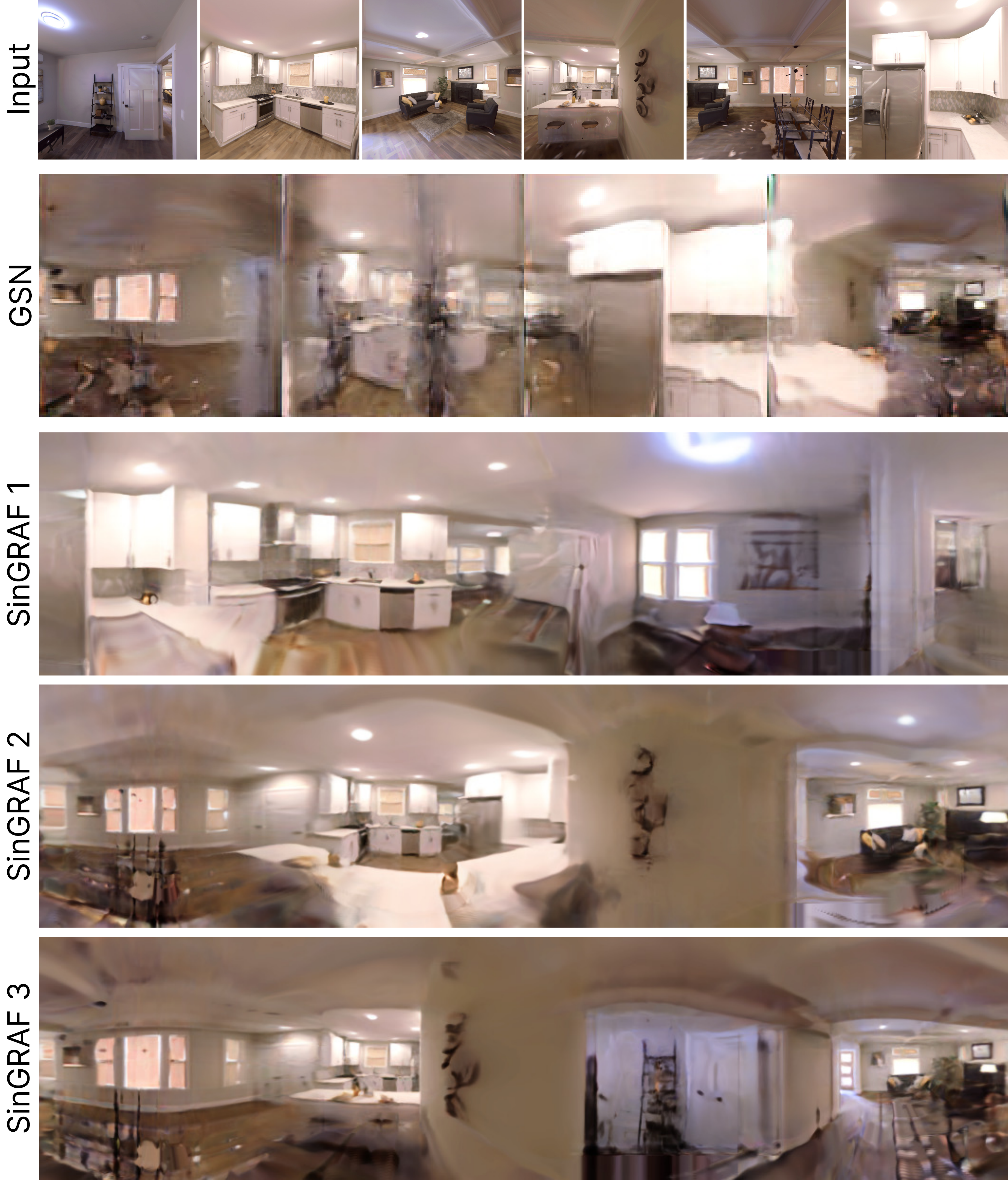}
        \caption{``apartment\_0'' scene from the Replica dataset.}
        \label{fig:results_apartment_0}
    \end{subfigure}
    \\
    \begin{subfigure}[b]{0.49\textwidth}
        \includegraphics[width=\linewidth]{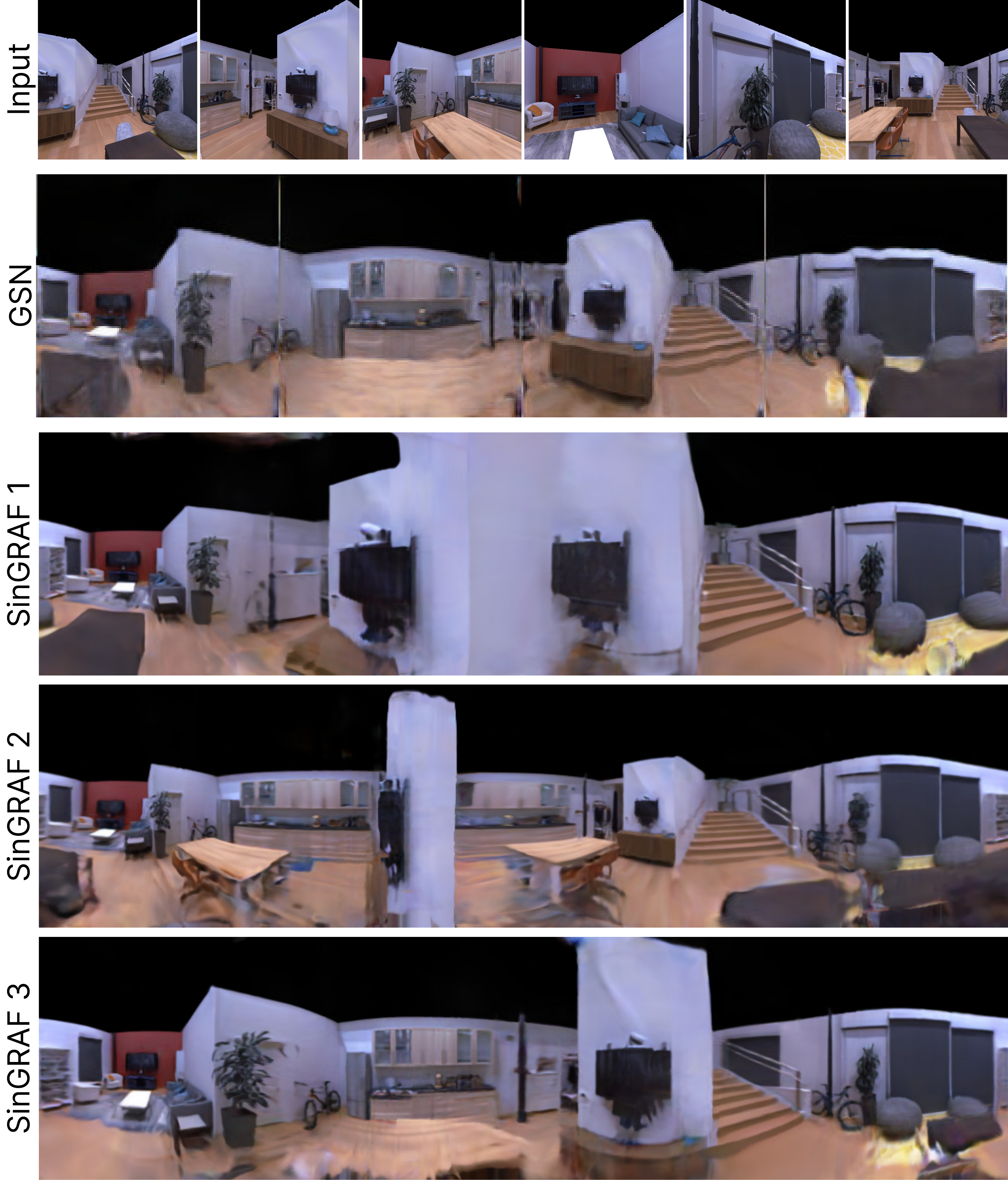}
        \caption{``frl\_apartment\_4'' scene from the Replica dataset. }
        \label{fig:results_frl_apartment_4}
    \end{subfigure}
    \hfill
    \begin{subfigure}[b]{0.49\textwidth}
        \includegraphics[width=\linewidth]{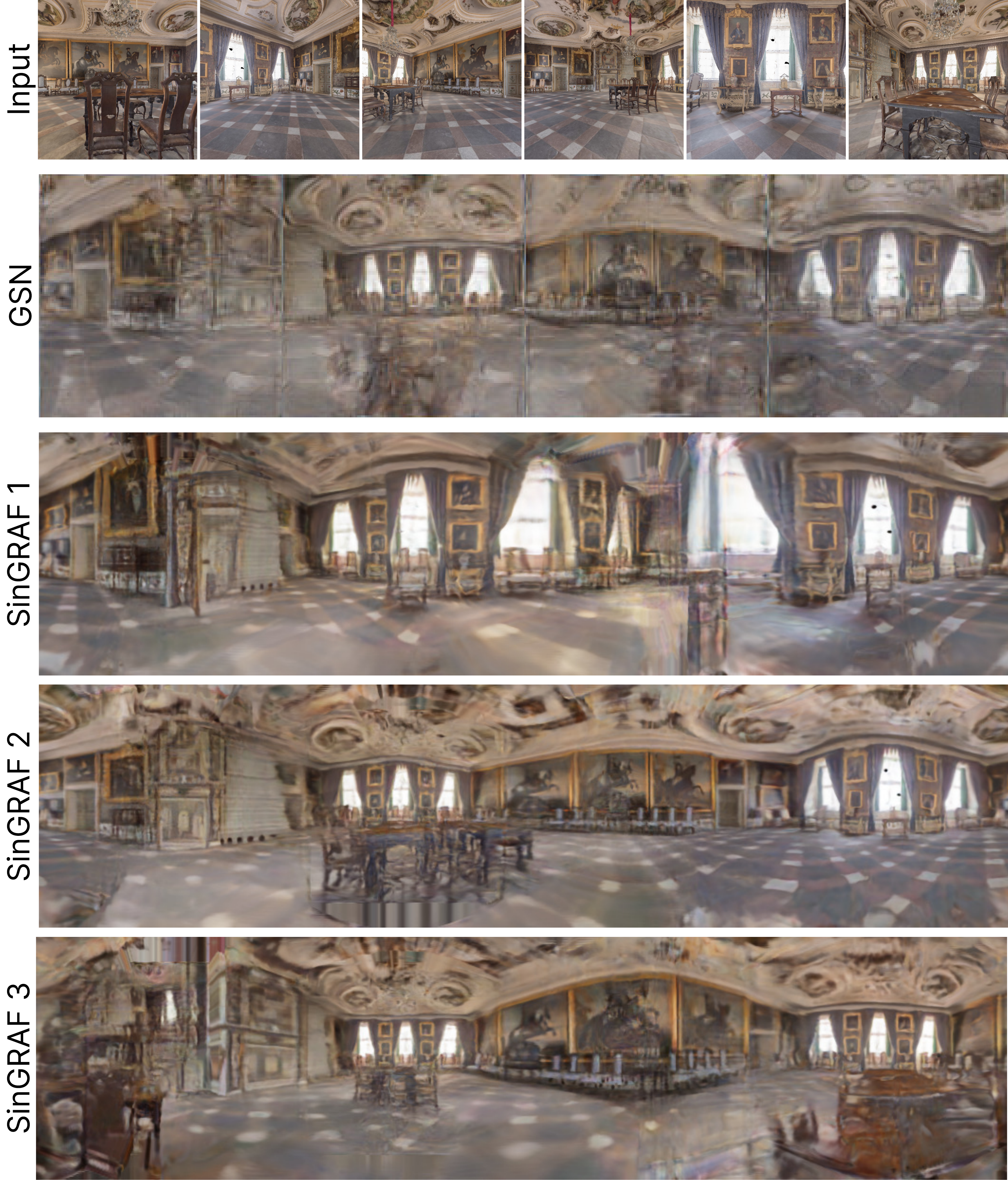}
        \caption{``castle'' scene from the Matterport3D dataset.}
        \label{fig:castle}
    \end{subfigure}
    \caption{Results of four difference scenes, as indicated. For each example, we show some of the input images, the result achieved by the GSN~\cite{DeVries_2021_ICCV} baseline, and three realizations of SinGRAF's results for the same input images. Note that GSN is mode-collapsed and not able to generate different realizations of this scene while SinGRAF is capable of generating a diverse set of realizations with high image quality.}
    \label{fig:results}
\end{figure*}

\subsection{Baselines}
As a baseline, we compare SinGRAF against the current state-of-the-art 3D scene generative method, GSN \cite{DeVries_2021_ICCV}. While numerous 3D-GAN methods were proposed, we could not find any other suitable baseline that demonstrated the modeling of apartment-scale scenes, such as the ones from the Replica dataset. Concurrently to our work, GAUDI \cite{bautista2022gaudi} extended GSN to model even larger-scale scenes with generative modeling, but the code is not publicly available for us to run. To compare against GSN, we use their public codebase and run their algorithm using the same 100-image datasets. During training, we sample the virtual cameras 
from the 10,000 camera samples used in their original paper.
Following the original implementation, the images are volume-rendered at $64\times 64$ resolution and upsampled to $128\times 128$ with a learned CNN.

\subsection{Metrics}
To measure the generated image quality we use the popular Kernel Inception Distance (KID) score, which reliably measures the distance between two sparsely sampled (N$\leq$500) image distributions. This is in contrast with FID, which introduces significant biases in the low-data regime. We randomly sample 500 images for all methods in the same way as each of them samples during training. The 500 ground truth images are sampled as we generate the training images.

We measure the diversity of scene generation by sampling multiple images from a fixed camera and computing average LPIPS distance (average pair-wise LPIPS distances), following \cite{Huang_2018_ECCV}. Note that, however, because our generated scenes are 3D, there exist loopholes to our diversity metric. For example, the diverse images generated from a fixed camera could be identical in the 3D space under rigid transformations. However,  we did not observe this edge case and observe that the high diversity score from a fixed view leads to diversity beyond rigid transformations and vice versa.

\begin{table}[t!]
    \centering
    \footnotesize
    \begin{tabular}{lcccc|cc}
        \toprule
         & \multicolumn{2}{c}{GSN ($128^2$)}
         & \multicolumn{2}{c|}{$\!\!\!$SinGRAF ($128^2$)}
         & \multicolumn{2}{c}{$\!\!$SinGRAF ($256^2$)} \\
         & KID$\downarrow$ & Div.$\uparrow$
         & KID$\downarrow$ & Div.$\uparrow$
         & KID$\downarrow$ & Div.$\uparrow$ \\
        \midrule
        office\_3   & .061 & .001 & \bf .044 & \bf .297 & .050 & .378 \\
        hotel\_0    & .049 & .012 & \bf .037 & \bf .413 & .046 & .490 \\
        apt.0       & .069 & .001 & \bf .037 & \bf .401 & .049 & .467 \\
        frl\_apt.4  & .052 & .001 & \bf .037 & \bf .335 & .055 & .408 \\ 
        castle      & \bf .050 & .001 & .064 & \bf .248 & .088 & .318 \\
        office\_0    & .075 & .001 & \bf .053 & .001 & .062 & .003 \\
        dynamic $\!\!\!\!$    & .089 & .013 & \bf .033 & \bf .298 & .050 & .365 \\
        \bottomrule
    \end{tabular}
    \caption{Quantitative Results. We measure the realism and diversity of the 3D scenes generated from SinGRAF and GSN on Replica and Matterport3D scenes. KID compares the distributional difference between the rendered and ground truth images of the scenes. We measure the diversity by rendering images with various latent vectors from fixed camera. Overall, we outperform the GSN baseline on both metrics in all but one case. }
    \label{tbl:comparison}
\end{table}

\subsection{Scene Generation Results}
We showcase the visual quality and diversity of our trained generative model across scenes. As can be seen in Figs.~\ref{fig:teaser},~\ref{fig:results_hotel_0},~\ref{fig:results_apartment_0},~\ref{fig:results_frl_apartment_4},  our method is able to synthesize plausible and realistic variants of the original scenes under a wide range of indoor scene environments. For example, the ``office\_3" scene shown in Fig.~\ref{fig:teaser} contains a meeting table (orange), a sofa (grey), and a coffee table (white). Note how our model is able to augment the room by duplicating, elongating, or rotating the meeting table and the sofa. For the ``hotel\_0" and ``apartment\_0" scenes shown in Figs.~\ref{fig:results_hotel_0},~\ref{fig:results_apartment_0}, our model was able to capture {\em structural} diversity of the scenes while preserving the details and general appearance of the scene. As shown in the results of ``frl\_apartment\_4" (Fig.~\ref{fig:results_frl_apartment_4}) our model was able to generate scenes with very different sizes by duplicating large scene structures. The results in the ``castle" scene, shown in Fig.~\ref{fig:castle}, features various structural changes along with the diversity of the number and locations of the chairs in the scene. Overall, SinGRAF was able to synthesize a remarkable amount of variation across scenes, even when the scene is small and simple, e.g., ``office\_3" or ``hotel\_0."

\begin{figure}[t!]
    \centering
    \includegraphics[width=.47\textwidth]{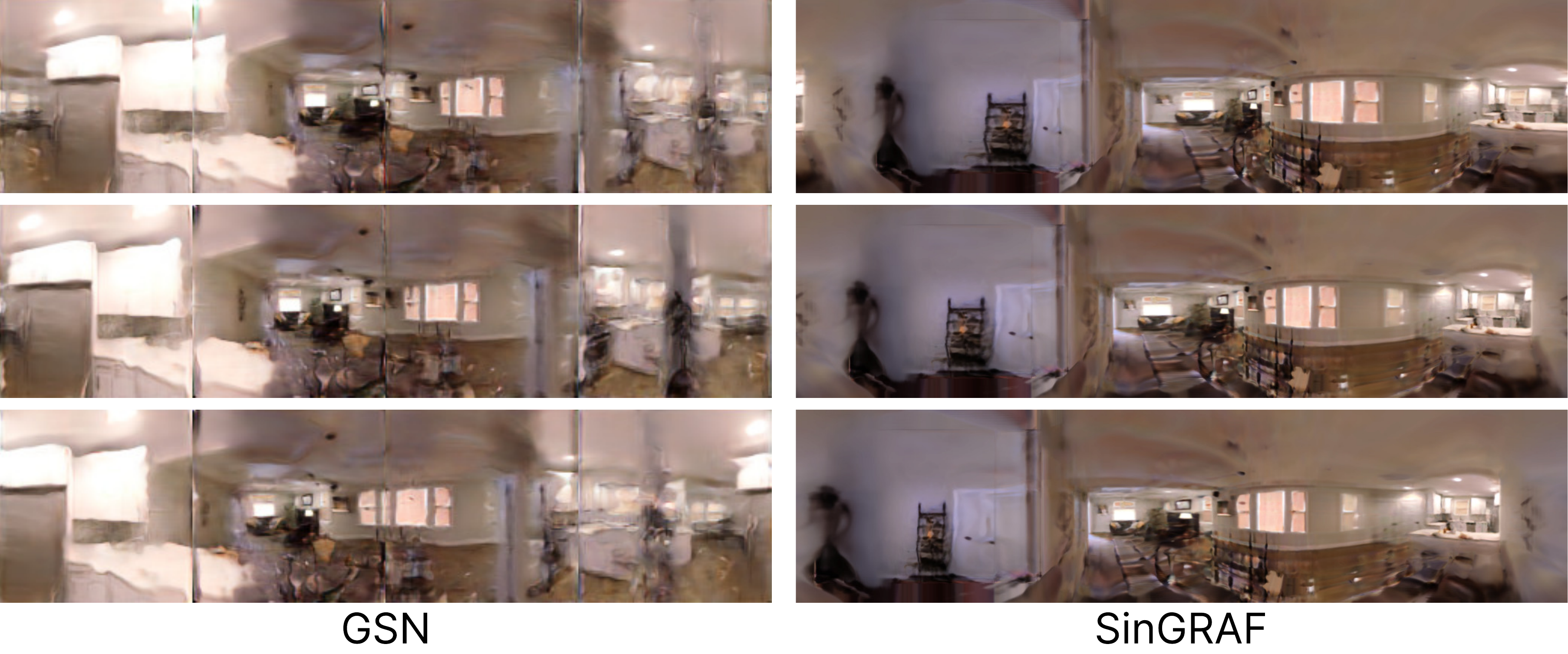}\\
    \includegraphics[width=.47\textwidth]{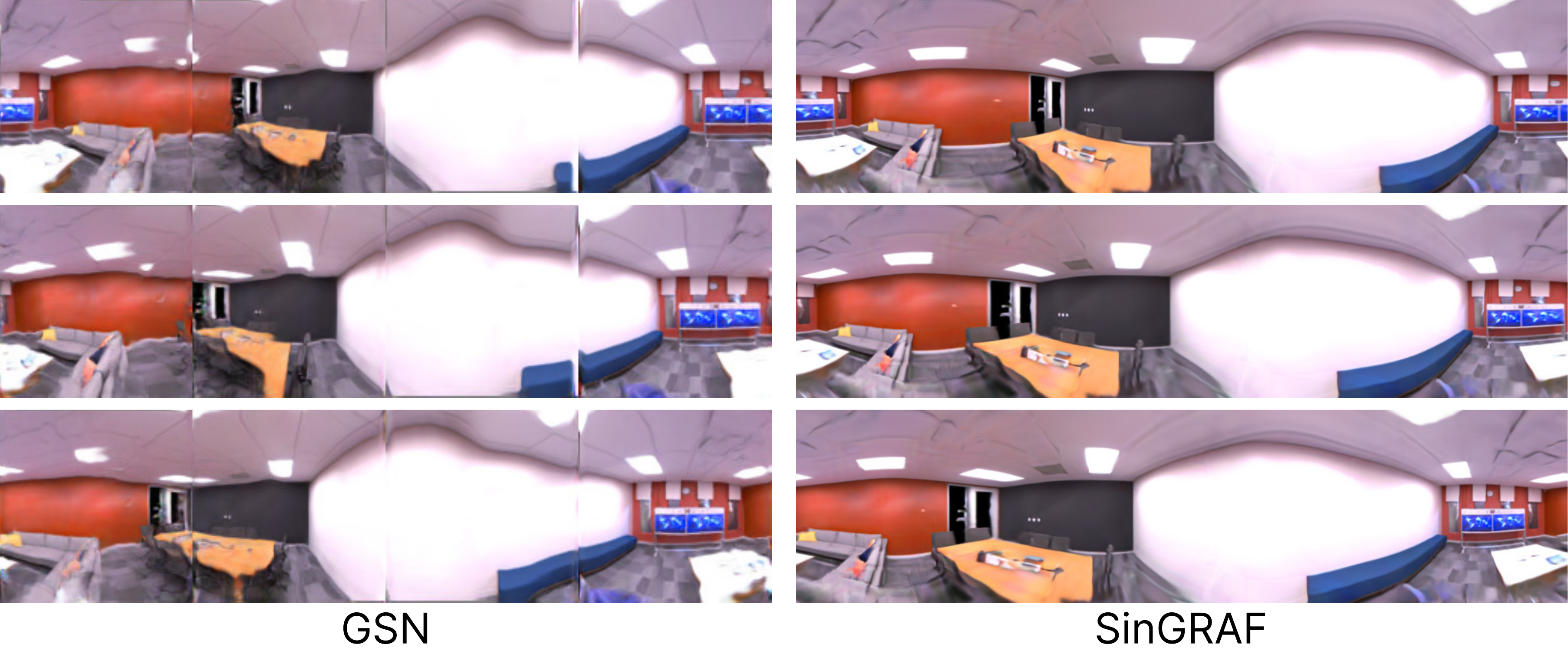}\\
    \caption{View consistency test with small rotations ($15^\circ$). We check the 3D consistency of the neural fields generated by SinGRAF and GSN by visualizing the scene from varying viewpoints. GSN produces inconsistent discontinuities and abrupt structure changes (i.e., plant in the right), while SinGRAF demonstrates smooth and 3D-consistent view changes. Note that the shown panorama naturally induces some image distortions. 
    }
    \label{fig:results_view}
\end{figure}

On the other hand, applying the strongest current baseline for scene generation, GSN \cite{DeVries_2021_ICCV}, produces mode-collapsed results without diversity for all of the tested scenes. Quantitatively, as shown in Tab.~\ref{tbl:comparison}, SinGRAF outperforms GSN both in terms of realism and diversity. 

\paragraph{Video Results.}
We urge readers to watch our supplementary videos to fully appreciate the quality and diversity of our 3D scenes.

\subsection{View-Consistent Scene Generation}
We notice that even for the single-mode results of GSN, the scene renderings are of suboptimal quality, containing spurious blur (Fig.~\ref{fig:results_apartment_0}) or unnaturally abrupt content changes by viewpoints (Fig.~\ref{fig:results_hotel_0}). We hypothesize that GSN learned to generate plausible images but does not learn to create consistent 3D structures. To test this hypothesis, we visualize both GSN and our method when rotating the panorama twice by 15$^\circ$ each (Fig.~\ref{fig:results_view}). For the case of GSN, the slight change of viewpoints resulted in significant structural changes, for example changing the room or changing the shape of the furniture. In contrast, SinGRAF reliably generates consistent images and scene structures with varying viewpoints, indicating strong 3D awareness of the learned representations.

\subsection{Modeling Scene Dynamics}
\begin{figure}[t!]
    \centering
    \includegraphics[width=.47\textwidth]{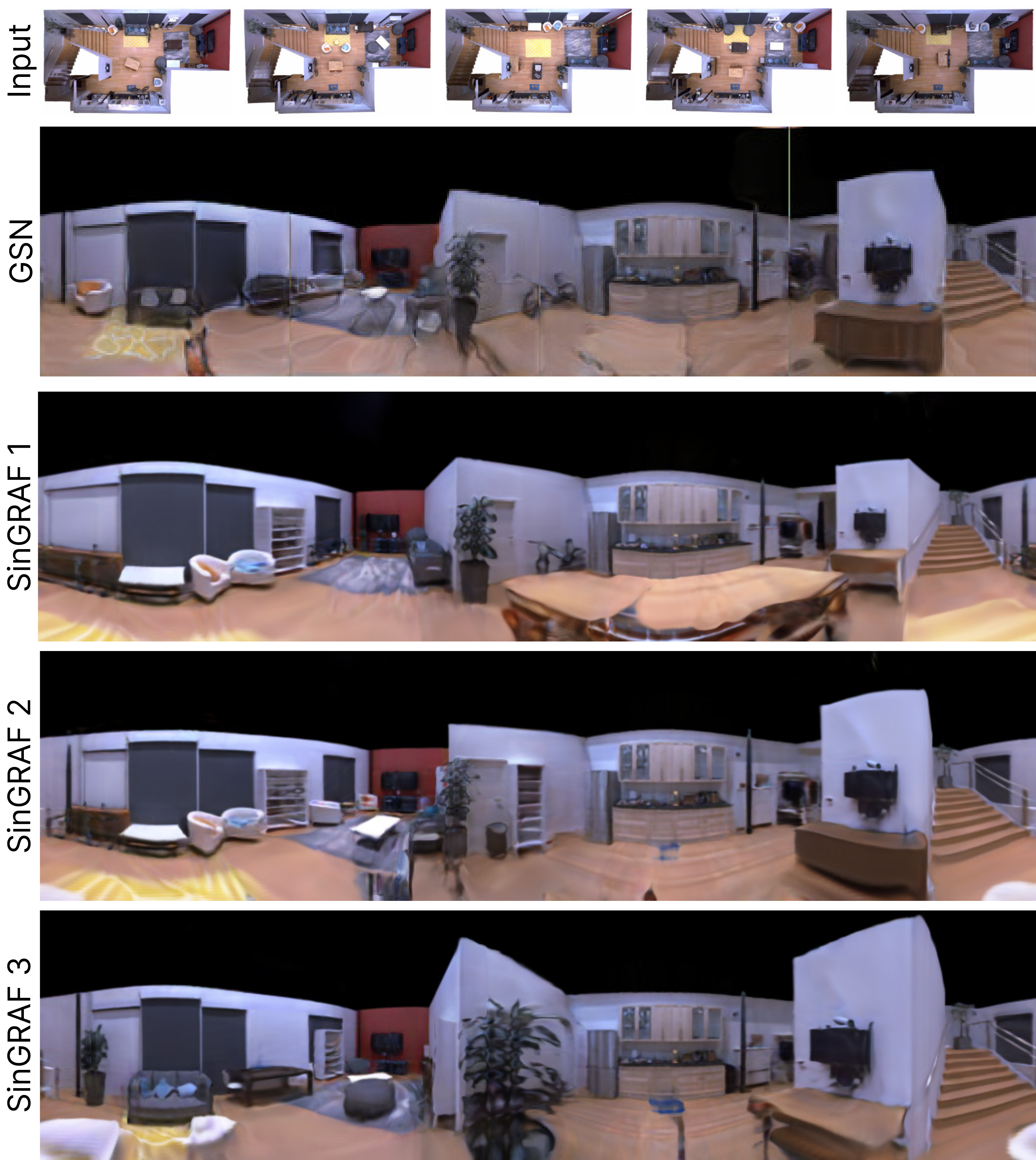}\\
    \vspace{-0.1em}
    \caption{Modeling scene dynamics. We train our model using 500 images from five different scene configurations of the ``frl\_apartment''. Our resulting generative model produces 3D scenes with highly diverse variations of objects and structures, e.g., the locations of furniture.}
    \label{fig:results_frl_apartment_dynamic}
\end{figure}

We highlight that SinGRAF is especially robust in capturing scene dynamics since it does not rely on any pixel-wise reconstruction loss. To verify this claim, we run our method on the ``frl\_apartment" scene in the Replica dataset that has been captured with various configurations (5 different scene setups). The synthesis  results shown in Fig.~\ref{fig:results_frl_apartment_dynamic} indeed confirm SinGRAF's ability to train a high-quality generative model on scenes that are not static. Note that, given diverse input configurations, SinGRAF was able to induce more variations to the scenes, by reorganizing the objects in the scenes.  We note that training on these dynamic configurations did not result in diverse scene generation of GSN -- it produced single mode outputs. 

\begin{figure}[t!]
    \centering
    \includegraphics[width=.47\textwidth]{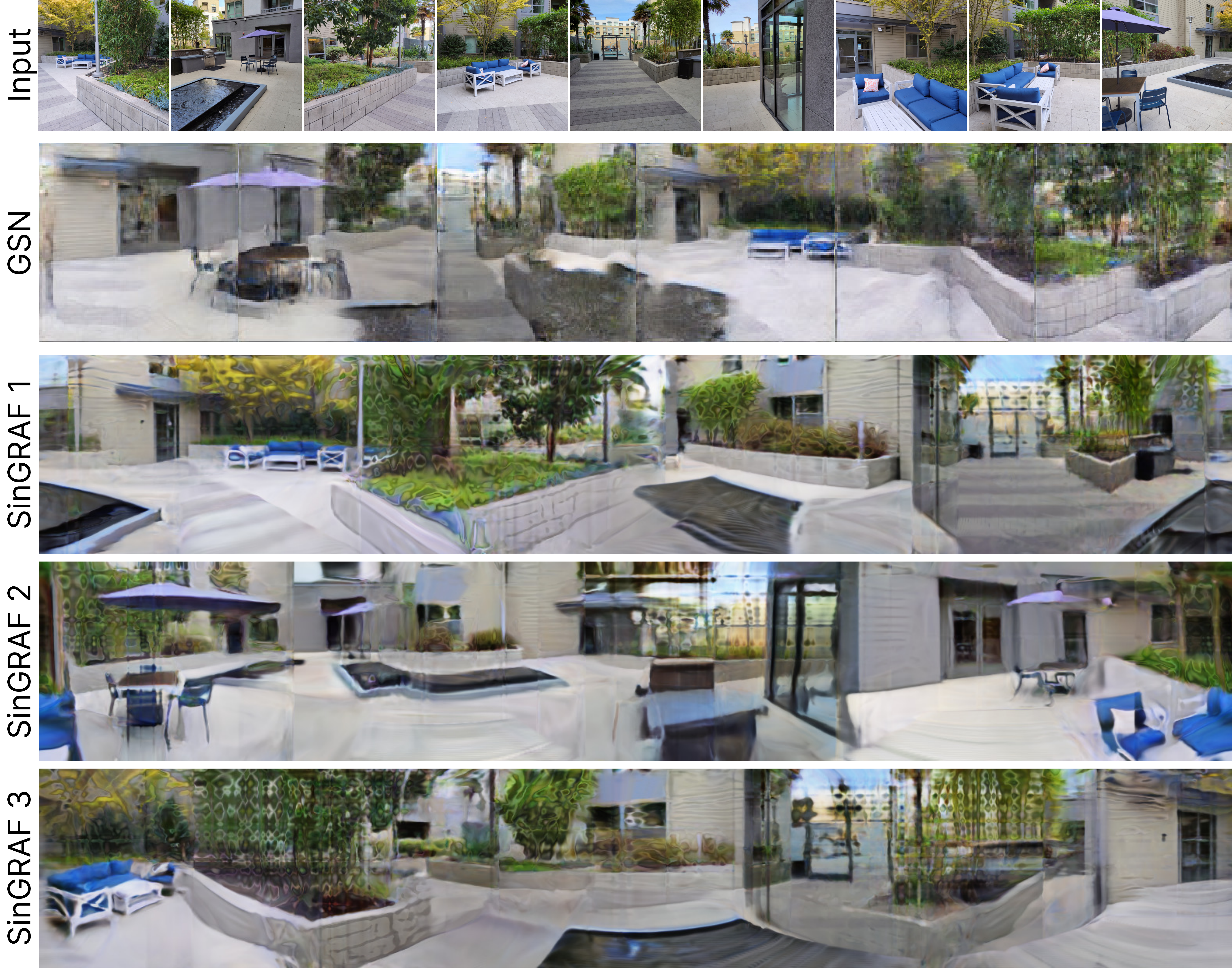}\\
    \vspace{-0.1em}
    \caption{Casually-captured scene. Given 100 input images photographed with a smartphone, our method is successful in generating high-quality variations of the 3D scene.
    }
    \label{fig:results_capture}
\end{figure}

\subsection{Towards Casually-captured Scenes}

We also test our method ``in the wild'', i.e., with captured content rather than pre-scanned scenes. For this experiment, we took 100 photographs of an outdoor apartment setting using a consumer-level smartphone. This scene is particularly challenging because it contains a lot of high-frequency textures, such as trees and grass, and a large dark window with view-dependent reflections. 
The camera setting also makes the problem difficult, as the intrinsic parameters are unknown and training images contain lens distortions.
We approximate the field of view of the cameras with $65^\circ$, and our model successfully generates variations with visually pleasing quality, as shown in Fig.~\ref{fig:results_capture}.
The average LPIPS distance is $0.001$ for GSN, and $0.372$ and $0.444$ for SinGRAF at $128^2$ and $256^2$ image resolution, respectively.
Although a KID score is not available for this example, because we only have 100 training images and no ground truth, this experiment demonstrates the potential of SinGRAF to be applied in the wild.

\subsection{Analysis}
\label{sec:analysis}
We analyze the effects of our design decisions through ablation studies next. The quantitative ablation results are displayed in Tab.~\ref{tbl:ablation}.

\begin{table}[t!]
    \centering
    \small
    \begin{tabular}{lcc|cc}
        \toprule
         & \multicolumn{2}{c|}{$128 \times 128$}
         & \multicolumn{2}{c} {$256 \times 256$} \\
         & KID$\downarrow$ & Div.$\uparrow$
         & KID$\downarrow$ & Div.$\uparrow$ \\
        \midrule
        full \& half-scale patches    & .183 & .001 & NA & NA \\
        progressive patches           & .046 & .308 & .068 & .374 \\
        + camera opt.               & .037 & .295 & .056 & .368 \\
        + perspective aug.          & .037 & .335 & .055 & .408 \\
        \bottomrule
    \end{tabular}
    \caption{Ablation study. We ablate our model to study the effects of patch discrimination, camera pose optimization, and perspective augmentation using the ``frl\_apartment\_4'' scene. Note that progressive patch scaling is essential for obtaining diverse scenes. Camera distribution optimization improves image quality while perspective augmentation maximizes scene diversity.}
    \label{tbl:ablation}
\end{table}

\paragraph{Patch Discrimination.}
As suggested by prior works on single-image GAN, learning the scene-internal patch statistics is crucial in inducing  diversity into generated results. Indeed, without the patch discrimination, i.e., using full-scale ($s=1.0$) images for adversarial training, results in a complete mode collapse of the model. Even when combining half-scale ($s=0.5$) and full-sale ($s=1.0$) patches as a 50\% mix, the training results in no diversity (first row of Tab.~\ref{tbl:ablation}), suggesting the importance of our progressive patch scaling strategy.

\paragraph{Camera Pose Optimization.}
We test the importance of our non-parametric camera distribution optimization scheme described in Sec.~\ref{sec: training}. As expected, adjusting the camera distributions result in higher realism of the generated outputs, resulting in lower KID scores (third row of Tab.~\ref{tbl:ablation}). On the other hand, the optimization scheme only slightly hurts the diversity of the generation.  

\paragraph{Perspective Augmentation.}
Given our low-data regime, a possible way of maximizing the diversity is via data augmentation strategies, e.g., perspective image augmentation (see Sec.~\ref{sec: training}). As expected, applying perspective image augmentation leads to higher diversity (fourth row of Tab.~\ref{tbl:ablation}). What we found interesting is that adding perspective augmentation of 15$^\circ$ did not result in lower KID scores. This shows that perspective augmentation is an effective strategy in our setting.

\begin{figure}[htbp]
    \centering
    \includegraphics[width=.47\textwidth]{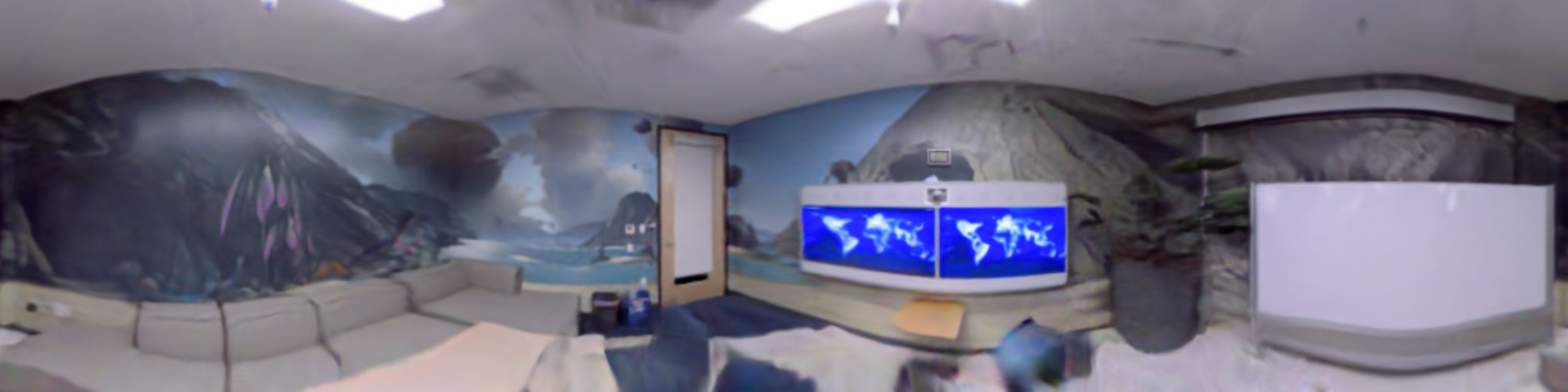}\\
    \vspace{-0.2em}
    \caption{Failure case of the ``office\_0'' scene. The lack of diversity for this scene is likely due to the paintings on the walls that uniquely determine the relative locations of most patches.}
    \label{fig:failure}
\end{figure}

\paragraph{Failure cases.}
We notice that for the ``office\_0" scene, shown in Fig.~\ref{fig:failure}, SinGRAF mode-collapses and fails to generate diversity. We believe that this occurs because the detailed paintings on the walls uniquely identify the location of the patches in relation to others. 

%% file: camera/sec6.tex
\section{Discussion}
\label{sec:discussion}

In this work, we take first steps towards a 3D generative model for a single scene from a set of unposed images. Our model, SinGRAF, is able to generate diverse and realistic variations of the given scene while preserving the semantics of human-made structures. In contrast to 2D single-image GAN methods \cite{Shaham_2019_ICCV}, which typically train a pyramid of generators, SinGRAF is trained with a single generator architecture, and is thus simple to implement and train. We achieve this by continuously adjusting the patch scales during training, leveraging the continuous neural fields representation. Given the unposed and scarce nature of our image data, we find it useful to optimize the camera pose distributions and apply perspective augmentations to the images. 

Below we discuss three interesting implications of our approach to the field of 3D computer vision.

\vspace{-1em}
\paragraph{Reconstructing Variations.} SinGRAF, in a sense, reconstructs the distribution of plausible 3D scenes given a set of images. Traditionally, the 3D vision community largely focused on finding the {\em most likely} mode of the 3D scene for reconstruction. Relaxing this constraint to instead sample from the posterior distribution of scenes could lead to new technologies or applications.

 \vspace{-1em}
\paragraph{Unposed Reconstruction.} SinGRAF does not use estimated poses of the input images as it relies on an adversarial rather than a pixel-wise loss. 
While SinGRAF, in its current form, is not designed to accurately reproduce the ground truth 3D scene, it might be extended to provide control of the {\em narrowness} of the distribution, depending on the applications. 

\vspace{-1em}
\paragraph{Dynamic Scene Reconstruction.} In contrast to traditional 3D reconstruction methods, scene dynamics would likely {\em improve} the quality and diversity of SinGRAF training (see Fig.~\ref{fig:results_frl_apartment_dynamic}). Such a trend implies that it could be easier to extend SinGRAF to operate on highly dynamics scenes, e.g., rock concerts, rather than developing physical models to handle complex scene dynamics.

\vspace{-1em}
\paragraph{Limitations.}
The quality and diversity of the SinGRAF outputs depend on the input images, scenes, and choice of views, which are hard to predict or control. Moreover, SinGRAF training is currently expensive and takes 1--2 days per scene with a single RTX 6000 GPU, although it is comparable to existing 3D GANs (e.g., GSN). We expect improvements in the training speed with the development of more efficient continuous representations, such as~\cite{muller2022instant,chen2022tensorf}.

%% file: camera/ack.tex
\section*{Acknowledgements}

This project was in part supported by Samsung, Stanford HAI, a PECASE from the ARO,
ARL grant W911NF-21-2-0104, a Vannevar Bush Faculty Fellowship, and
gifts from the Adobe and Snap Corporations.

%% file: camera/supp1.tex
\section{Video Results}
We highly encourage readers to view our supplementary video containing visualizations of our 3D scenes, latent interpolations, and comparisons against GSN. The video results are best suited for appreciating the 3D consistency, quality, and diversity of our generated scenes. The name of the attached video file is ``singraf\_video.mp4."

\section{Implementation Details}
\label{sec:implementation}

\subsection{Progressive Patch Scaling}

In practice, we used the input resolution of $512\times 512$ for the input images ($H'=512$) and used the fixed patch size of $64\times 64$ ($H=64$). When the scale $s=1$ the $64\times 64$ patch covers the entire image.

 To progressively scale down the patches, we reduce the patch scale $s$ for the first 100 epochs. Because we are using 1,000 sample batches for each epoch, we gradually reduce $s$ during the course of 100,000 iterations and then fix $s$ for the rest of the training. Typically our model exhibits the best KID score around 300 to 400 epochs. 

We randomly sample the scale factor $s$ independently for each image instance during training. We use a time-varying uniform distribution for the patch scale: $s\sim \texttt{U}(s_{\min}(t), s_{\max}(t)),$ where $t$ is the epoch index. In practice, we used $s_{\min}(0)=0.6$ and $s_{\max}(0)=0.8$ in the beginning of the training and $s_{\min}(100)=0.25$ and $s_{\max}(100)=0.55$ at epoch 100. $s_{\min}(t)$ and $s_{\max}(t)$ values are interpolated linearly as a function of $t$ over the course of 100 epochs.

\subsection{Data Augmentation}
\begin{figure}[htbp]
    \centering
    \includegraphics[width=.47\textwidth]{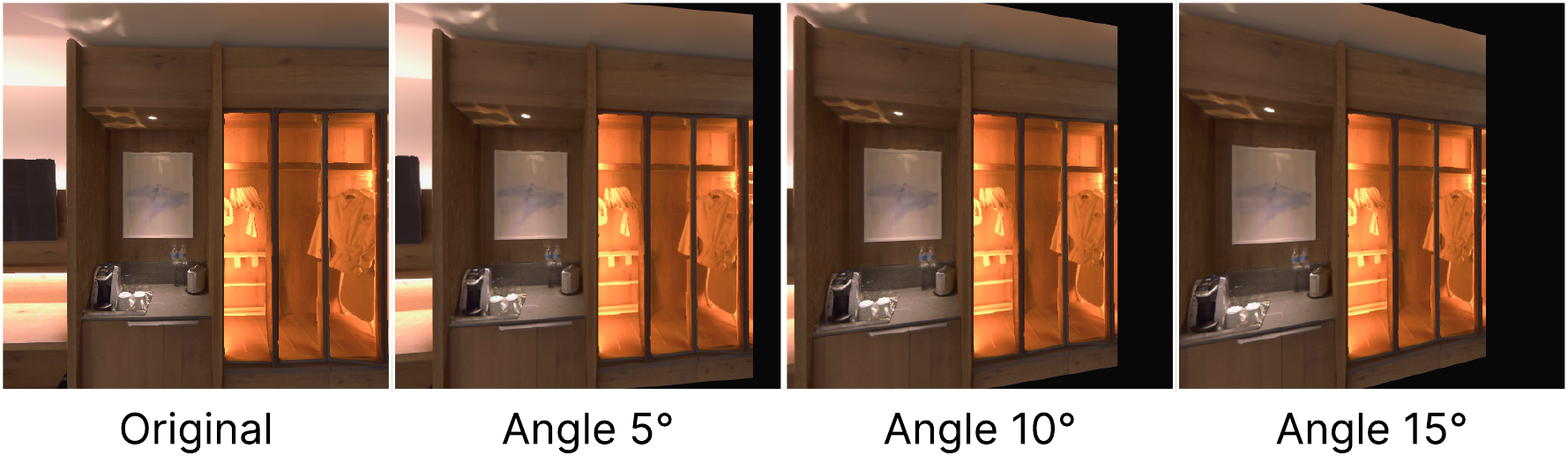}\\
    \includegraphics[width=.47\textwidth]{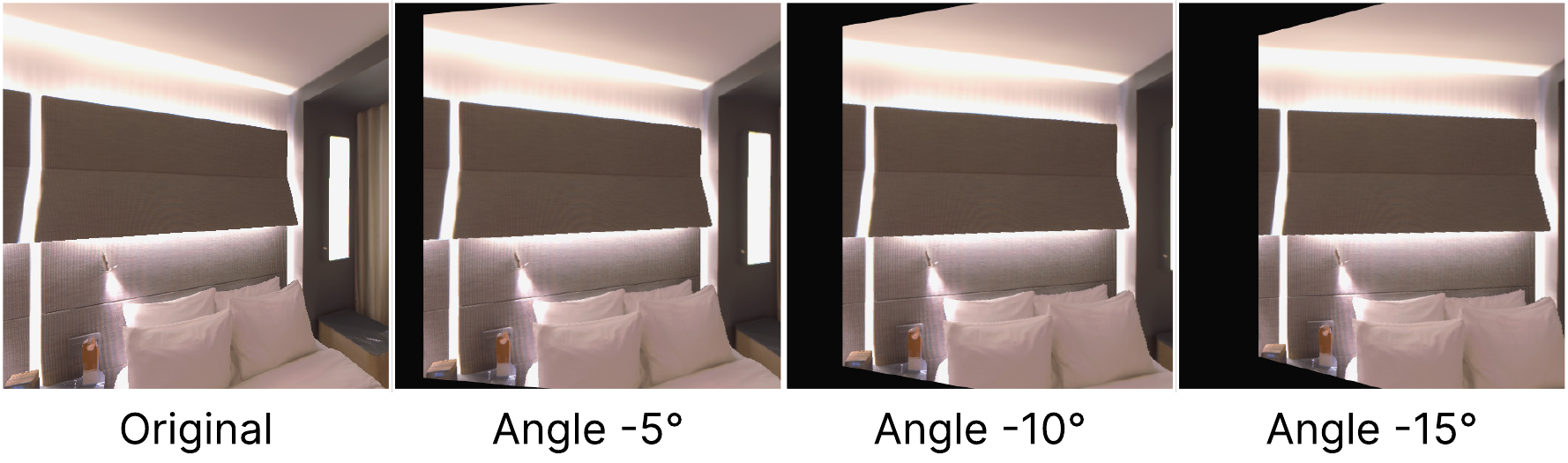}\\
    \caption{Visualizations of the perspective augmentation applied at different angles. During training, we apply up to $15^\circ$ perturbation to promote diversity.
    }
    \label{fig:perspective_augmentation}
\end{figure}

As described in the main text, we schedule the increase of angles used for perspective augmentation during training. Similar to the progressive patch scaling, we gradually linearly increase the maximum augmentation angle for 100 epochs. We start from the angle range of $[0^\circ, 0^\circ]$ to $[-15^\circ, 15^\circ]$ linearly over the course of 100 epochs. For each real image patch we randomly and independently sample an angle from the current range and apply the perspective augmentation on the height axis. 
An example visualization of this augmentation could be seen in Fig.~\ref{fig:perspective_augmentation}.

\subsection{Camera Pose Optimization}\label{sec: camera_opt}
As described in the main text, we non-parametrically optimize the camera pose distribution during training. Given a set of 1,000 camera poses: $\mathcal{T}=\{T_1,...,T_{1000}|T_i\in SE(3)\}$. During training, we randomly sample camera poses from $\mathcal{T}$ and render the scenes from the cameras. We let these camera poses as optimizable variables and backpropagate the gradients from the adversarial loss to optimize the individual poses. Optimizing for the $SE(3)$ transformation is known to be a difficult task, and directly optimizing the values of the transformation matrix is difficult because the matrix can escape the manifold of $SE(3)$, e.g., $RR^{\intercal}\neq I.$ Therefore, we decompose the matrix into multiple components for ease of optimization.

Specifically, given a transformation $T,$ we can decompose it into:
\begin{equation}
T=\begin{pmatrix}
R & p\\
0 & 1
\end{pmatrix}: \, R\in SO(3), \, p\in \mathbb{R}^3.
\end{equation}
The rotation matrix can be further decomposed into a multiplication of 3 matrices: 
\begin{equation}
    R=R_z R_y R_x: \, R_{\{\}}\in SO(3),
\end{equation}
where $R_z$, $R_y$, and $R_x$ are respectively rotation matrix about $z$, $y$, and $x$ axis. For example, we have that:
\begin{equation}
R_z = 
\begin{pmatrix}
\cos\theta_z & -\sin\theta_z & 0\\
\sin\theta_z & \cos\theta_z & 0 \\
0 & 0 & 1
\end{pmatrix}:\, \theta_z \in \mathbb{R},
\end{equation}
where $\theta_z$ is the rotation around the $z$ axis. While it is possible to parameterize our rotations with $\theta_z$, $\theta_y$, and $\theta_x$, directly optimizing for the Euler angles is known to be difficult \cite{zhou2019continuity}, as there is a discontinuity at $\theta=0$. Therefore, we parameterize each rotation with the cosine and sine of the Euler angle for each axis.
Let us denote $\mathcal{R}_z$ to be the data structure we carry for rotation about the $z$ axis:
\begin{equation}
    \mathcal{R}_z:=[\cos\theta_z,\sin\theta_z].
\end{equation}
We similarly define $\mathcal{R}_y$ and $\mathcal{R}_x$ using cosine and sine of the angles. Then, we know that we can uniquely construct the rotation matrix $R$ from the $\mathcal{R}$'s. For the whole transformation, we carry the four data structures to fully describe and construct the matrix $T$, which are: $[\mathcal{R}_z,\mathcal{R}_y,\mathcal{R}_x,p]$.
We can optimize for these variables during training by backpropagating the adversarial losses. Note that for each update of the rotation parameter $\mathcal{R}_{\{\}}$, we need to make sure that the cosines and sines are proper, by normalizing it so that $||\mathcal{R}_{\{\}}||=1.$
In practice, we assume that cameras are located at the same height and rotate along vertical axis. Thus, we only optimize $p_x$, $p_z$ and $\mathcal{R}_y$  during the early stages of training where the expected patch scale is larger than $0.5$.

\subsection{Training Details}

The balancing parameters for the regularization terms in Eq.(5) of the main paper are 
 $\lambda_{1} = 0.5$ and $\lambda_{2} = 50$. 
We use the spatial resolution of $256 \times 256$ and feature channel of $C=32$ for tri-plane representation~\cite{Chan2022}, which is generated by a StyleGAN2 \cite{karras2020analyzing} generator which is modulated by a noise vector $\mathbf{z} \sim \mathbb{R}^{128}$.
For rendering, we compute the feature of each sample along a ray via bilinear interpolation followed by concatenation (for the three planes), and process the feature using a decoder $\textrm{\sc{mlp}}$ to finally obtain color and density value.
The decoder $\textrm{\sc{mlp}}$ is composed of two shared linear layers, one layer for the density branch, and two additional layers for the RGB branch.
Each hidden layer uses 64 hidden units with leakyReLU activation except the final ones used for outputting density and RGB values.
For volume-rendering we used 
 $96$ samples per ray without importance sampling, to generate patches of $64 \times 64$ resolution.
 When applied the patch scale of $s=0.25$, the effective resolution of each patch is $256\times 256$, i.e., the amount of details that exists in each patch would be obtained when rendering at $256\times 256$. Therefore, we can render our models at  $256 \times 256$ resolution with details even though we trained our models with $64\times 64$ patches, without any 2D upsampling networks.
 As described in the main text, we adopt the StyleGAN2 discriminator architecture to process the patches, but we additionally concatenate the scale of each patch.
We will release the source code upon acceptance.

%% file: camera/supp3.tex
\section{Additional Analysis}
\label{sec:additional}

\subsection{Dataset Selection}
\begin{figure}[ht!]
    \centering
    \includegraphics[width=.47\textwidth]{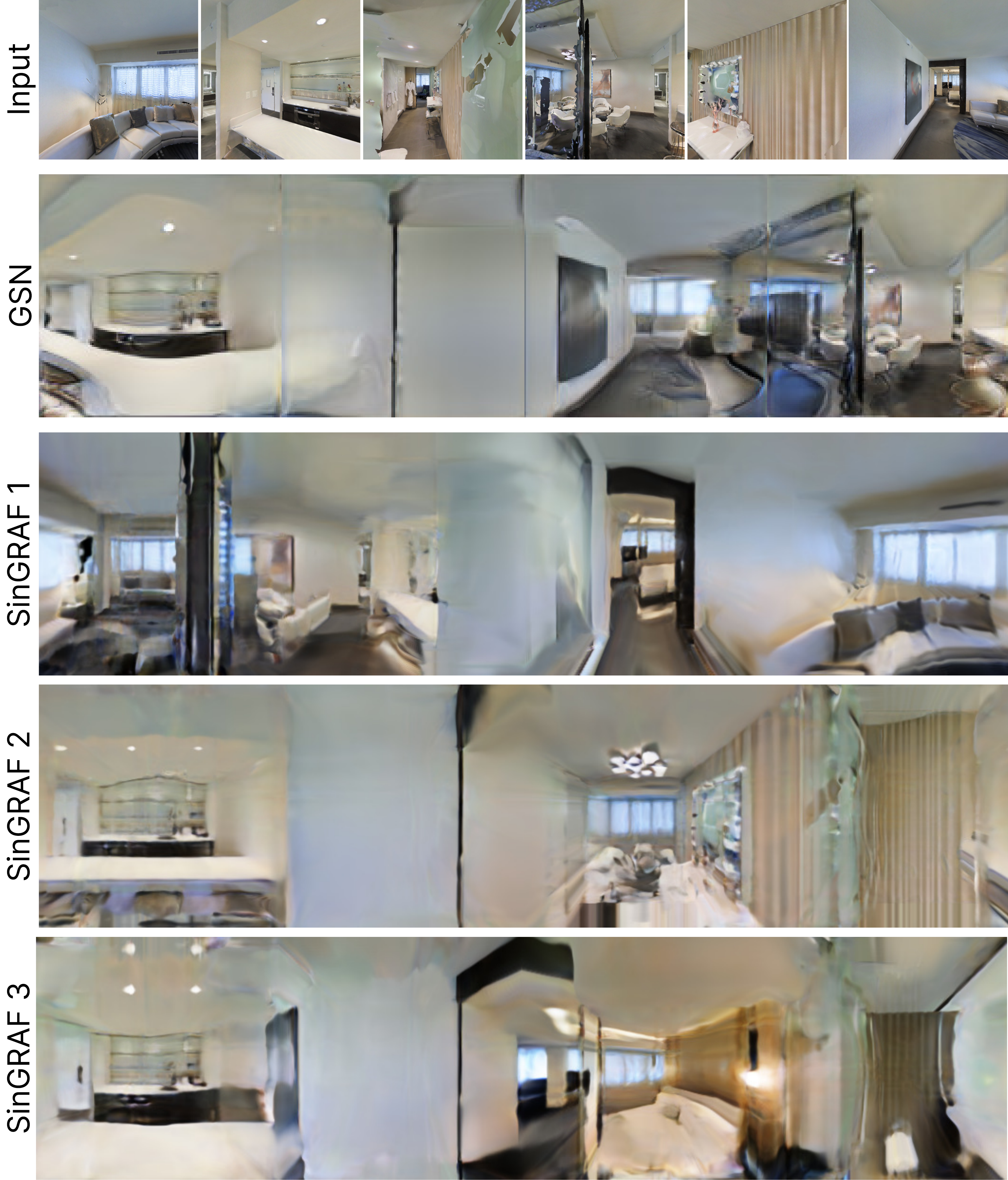}\\
    \caption{Additional Matterport3D scene.}
    \label{fig:newresults}
\end{figure}

Training 3D GANs, including GSN and \moniker{}, takes a long time. With the finite computational resources at our disposal, it was simply not possible to run \moniker{} on all scenes of a large 3D dataset such as Matterport3D. Therefore, we chose a representative subset of the Replica dataset and further stress-test our method on wildly different scene examples of a ballroom of Matterport3D and a custom-captured outdoor scene. We provide results on one additional Matterport3D scene (Fig.~\ref{fig:newresults}; \moniker{}: KID $\textbf{0.050}$ and Div. $\textbf{0.447}$; GSN: KID $0.087$ and Div. $0.001$). Similar to other scenes, GSN fails to produce diversity, while \moniker{} generates diverse and realistic scenes.

\subsection{Additional Empirical Analysis}
\paragraph{Number of input images}
We conduct an empirical study to gauge the effects of varying the number of input images. As shown in Tab.~\ref{tbl:num_images}, we observe that reducing the input image number degrades the quality and \moniker{} can generate diverse scenes for as few as 50 images. On the other hand, GSN fails to generate diversity for all ablated experiments due to mode collapse. 

\begin{table}[h]
    \centering
    \scriptsize
    \begin{tabular}{@{\hspace{-0.1em}}rcccc|cc|cc@{\hspace{-0.1em}}}
        \toprule
         & \multicolumn{2}{c}{GSN ($128^2$)}
         & \multicolumn{2}{c|}{$\!\!\!$SinGRAF ($128^2$)}
         & \multicolumn{2}{c}{$\!\!$SinGRAF ($256^2$)}
         & \multicolumn{2}{|c}{$\!\!$NeRF ($256^2$)$\!\!$}\\
         \# & KID$\downarrow$ & Div.$\uparrow$
         & KID$\downarrow$ & Div.$\uparrow$
         & KID$\downarrow$ & Div.$\uparrow$
         & KID$\downarrow$ & Div.$\uparrow$\\
        \midrule
        100 & .052 & .001 & \textbf{.037} & \textbf{.335} & \textbf{.055} & \textbf{.408} & .277 & 0.0 \\
        50  & .113 & .001 & \textbf{.071} & \textbf{.362} & \textbf{.104} & \textbf{.426} & .269 & 0.0 \\
        10  & .238 & .001 & \textbf{.130} & \textbf{.002} & \textbf{.159} & \textbf{.005} & .263 & 0.0 \\
        \bottomrule
    \end{tabular}
    \caption{Ablation over the number of input images using the ``frl\_apartment\_4" scene.
        }
    \label{tbl:num_images}
\end{table}

\paragraph{Comparison with NeRF variants}
While NeRF and their variants are used for obtaining 3D structure of a specific scene rather than generating diversity, we believe that 
NeRF variants are interesting baselines to put the realism of our generated scenes in context. 
To this end, we train Instant-NGP~\cite{muller2022instant} and measure KID of generated views at held-out poses on a Replica scene. To make the comparison fair, we do not use the ground truth camera poses but instead estimate the camera poses using an off-the-shelf structure-from-motion library, i.e., COLMAP \cite{schonberger2016structure}. As shown in the numerical results on Tab. \ref{tbl:num_images}, due to the small number of training views, the test view images of the NeRF variant are of very low quality.

\paragraph{Comparison with EG3D}
We trained a EG3D~\cite{Chan2022} baseline model for ``frl\_apartment\_4'' to show an additional comparison against a 3D-GAN algorithm. As expected, it fails to learn diversity (KID 0.078, Div. 0.008) without our continuous-scale patch discrimination.

\paragraph{Discriminator scale conditioning} As described in the main text, we condition our discriminator with the patch scale. We conduct an experiment to test of effect of the scale conditioning, which shows that without discriminator scale conditioning, the KID of the ablated \moniker{} on ``frl\_apartment\_4" is $0.070$ (worse quality) with a similar diversity score of $0.341$ compared to our model with the scale conditioning (KID $0.037$, Div. $0.335$).

\begin{figure}[ht]
    \centering
    \includegraphics[width=.47\textwidth]{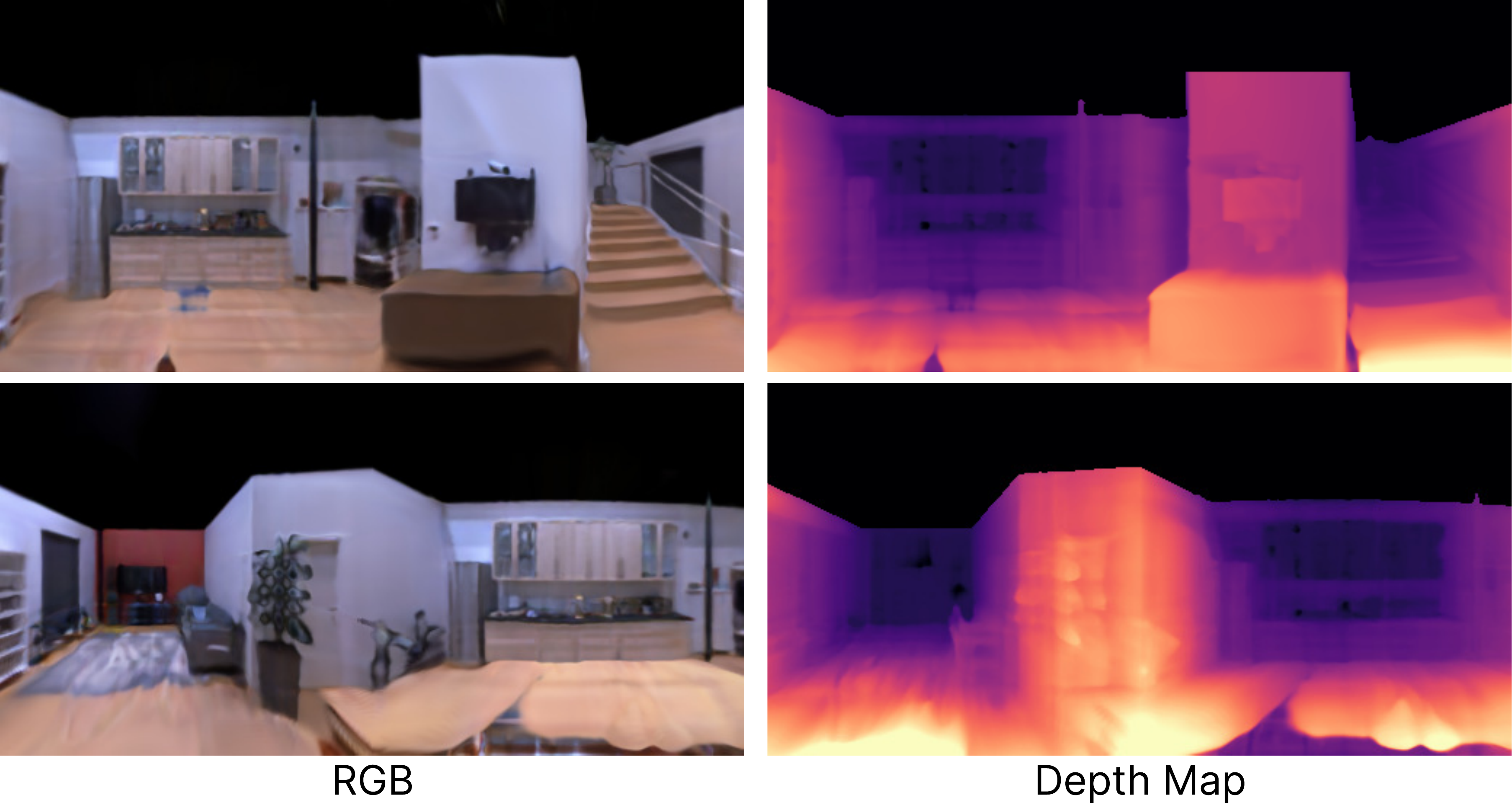}\\
    \caption{Depth map visualization (``frl\_apartment\_dynamic").}
    \label{fig:depth_map}
\end{figure}

\paragraph{Depth map visualization} To test the validty of the 3D structure generated by \moniker{}, we show example depth maps in Fig.~\ref{fig:depth_map}.

\subsection{Note on the GSN comparison}

We trained GSN~\cite{DeVries_2021_ICCV} using the exact same settings used in their paper and published code except that we did not use the depth maps. The random poses are sampled from a 10K pose basket and jittered.

To make the comparison fair against our setup using the 1K pose basket, we tried reducing the size of the random pose distribution from 10K to 1K for GSN on ``frl\_apartment\_4" but did not see any difference in quality or diversity (KID 0.055, Div. 0.001). Moreover, we tried GSN training with our pose sampling method, but it fails to converge (KID 0.599, Div. 0.0).
The total number of parameters of GSN (25.4M) is larger than ours (20.7M). We tried varying GSN parameters, discriminator resolution, and pose sampling, but couldn't achieve diversity.

%% file: camera/supp2.tex

\subsection{Visualization of the Diversity Metric}
\begin{figure}[htbp]
    \centering
    \includegraphics[width=.47\textwidth]{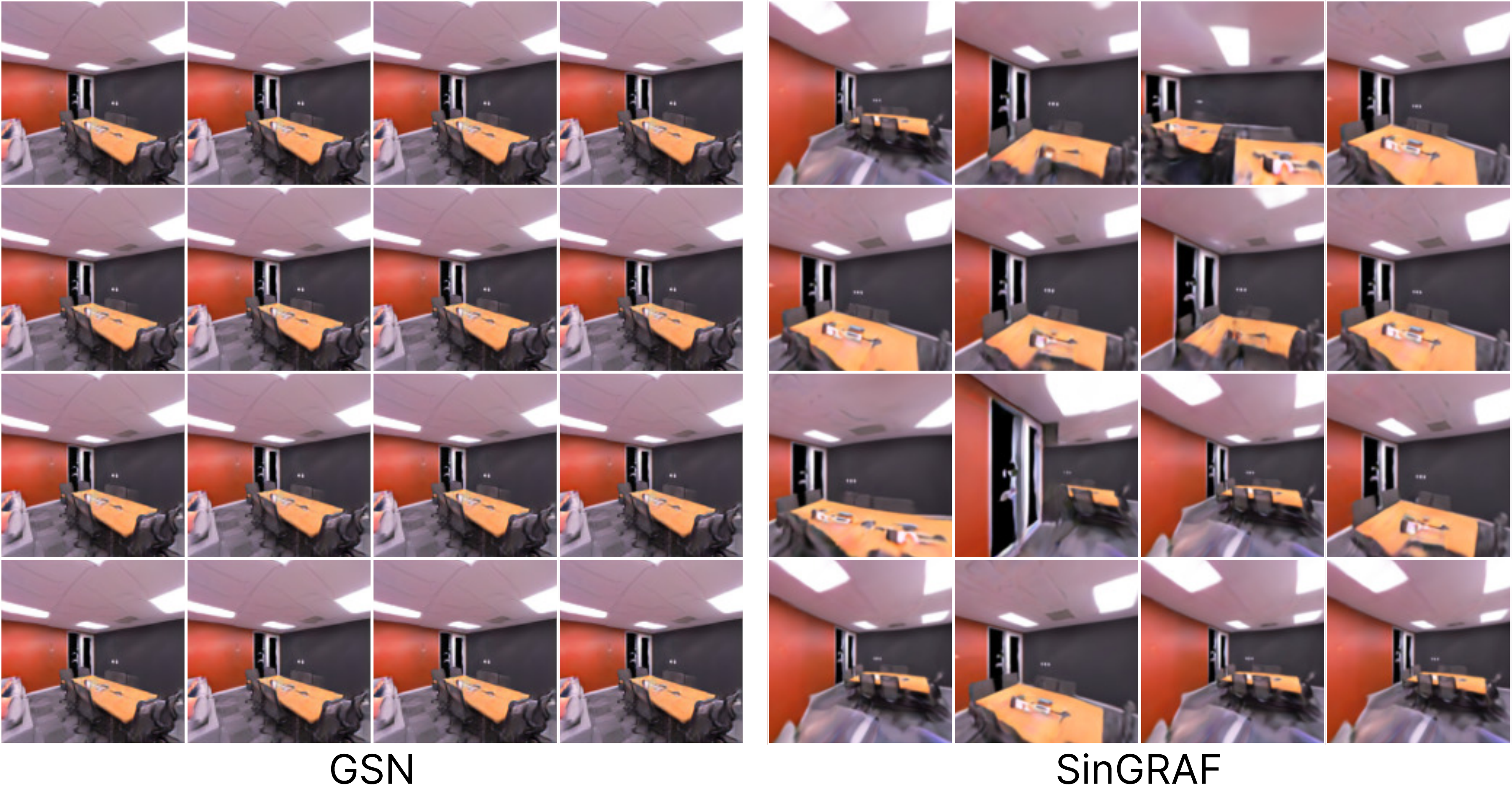}\\
    \caption{Visualizing randomly sampled 3D scenes from a fixed camera view. Notice how GSN's scenes do not change with the varying latent, indicating mode collapse, while SinGRAF presents highly diverse renderings.
    }
    \label{fig:fixed_view}
\end{figure}

As discussed in the main text, we measure the diversity of the 3D generative models by fixing a camera and rendering with randomly sampled latent codes. In Fig.~\ref{fig:fixed_view}, we show examples of such renderings. Note how we can tell that GSN's model has collapsed to a single mode, thus no variations from the fixed viewpoints. On the other hand, notice how SinGRAF generates highly diverse renderings of the same scenes from the fixed camera with varying latent.

\subsection{Latent Code Interpolation}
\begin{figure*}[t!]
    \centering
    \includegraphics[width=.99\textwidth]{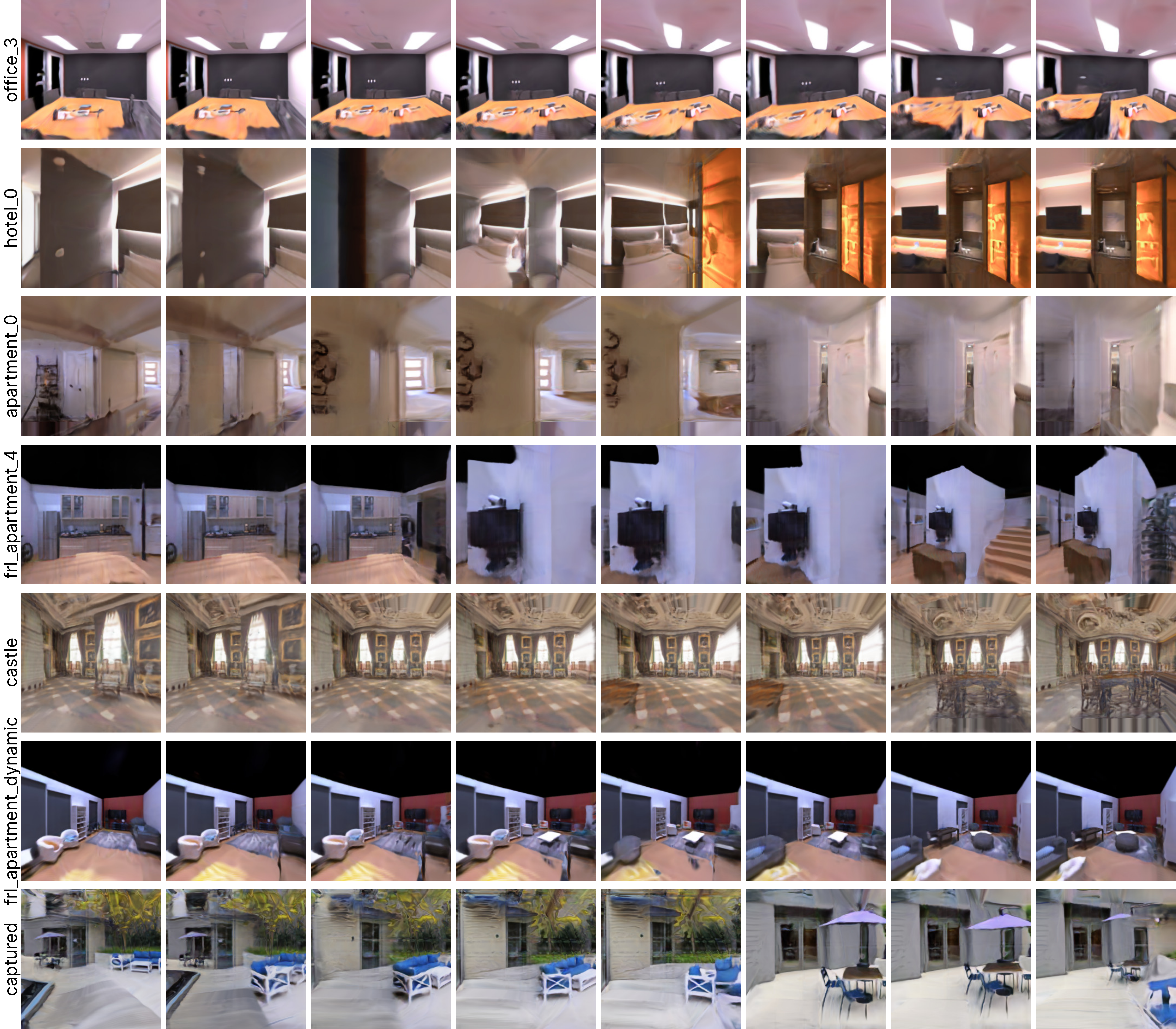}\\
    \caption{Latent code interpolations. We fix the camera viewpoint per scene and render the scene using latent vectors interpolating between two latent vectors, whose scenes are shown on the two extreme sides.
    }
    \label{fig:latent_interpolation}
\end{figure*}

To showcase the rich and smooth latent space we learn via training SinGRAF on single scenes, we visualize the latent space interpolation by fixing a camera and rendering the scene using latent vectors obtained via linear interpolation between two latent vectors. The results are shown in Fig.~\ref{fig:latent_interpolation}, demonstrating high-quality, and diverse latent embedding of the single scenes. We highly encourage readers to view our video results for animated interpolation in the latent space.

\subsection{Perspective Rendering Results}
\begin{figure*}[t!]
    \centering
    \includegraphics[width=.99\textwidth]{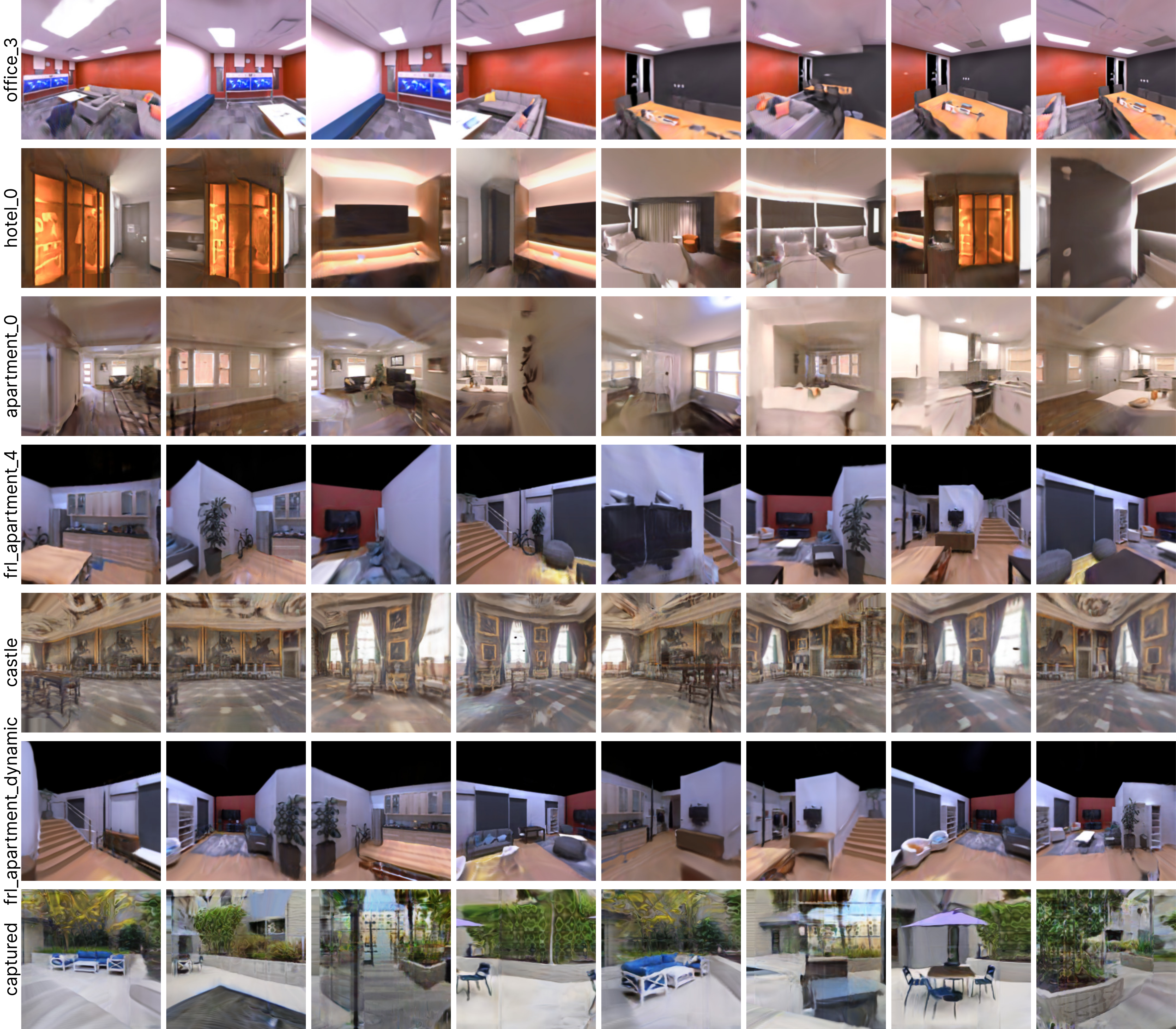}\\
    \caption{Perspective rendering results. We show perspective renderings of our trained scenes from randomly chosen latent codes and camera poses.
    }
    \label{fig:result_sample}
\end{figure*}

In the main text, we have only visualized our scenes in the panorama form. In Fig.~\ref{fig:result_sample}, we show renderings of the scenes from randomly chosen latent codes and camera poses using a perspective camera model. Note that we sampled the camera poses using the distribution $\mathcal{T}$, which is the result of the optimization process described in Sec.~\ref{sec: camera_opt} during training.

\subsection{Failure Case}
\begin{figure}[htbp]
    \centering
    \includegraphics[width=.47\textwidth]{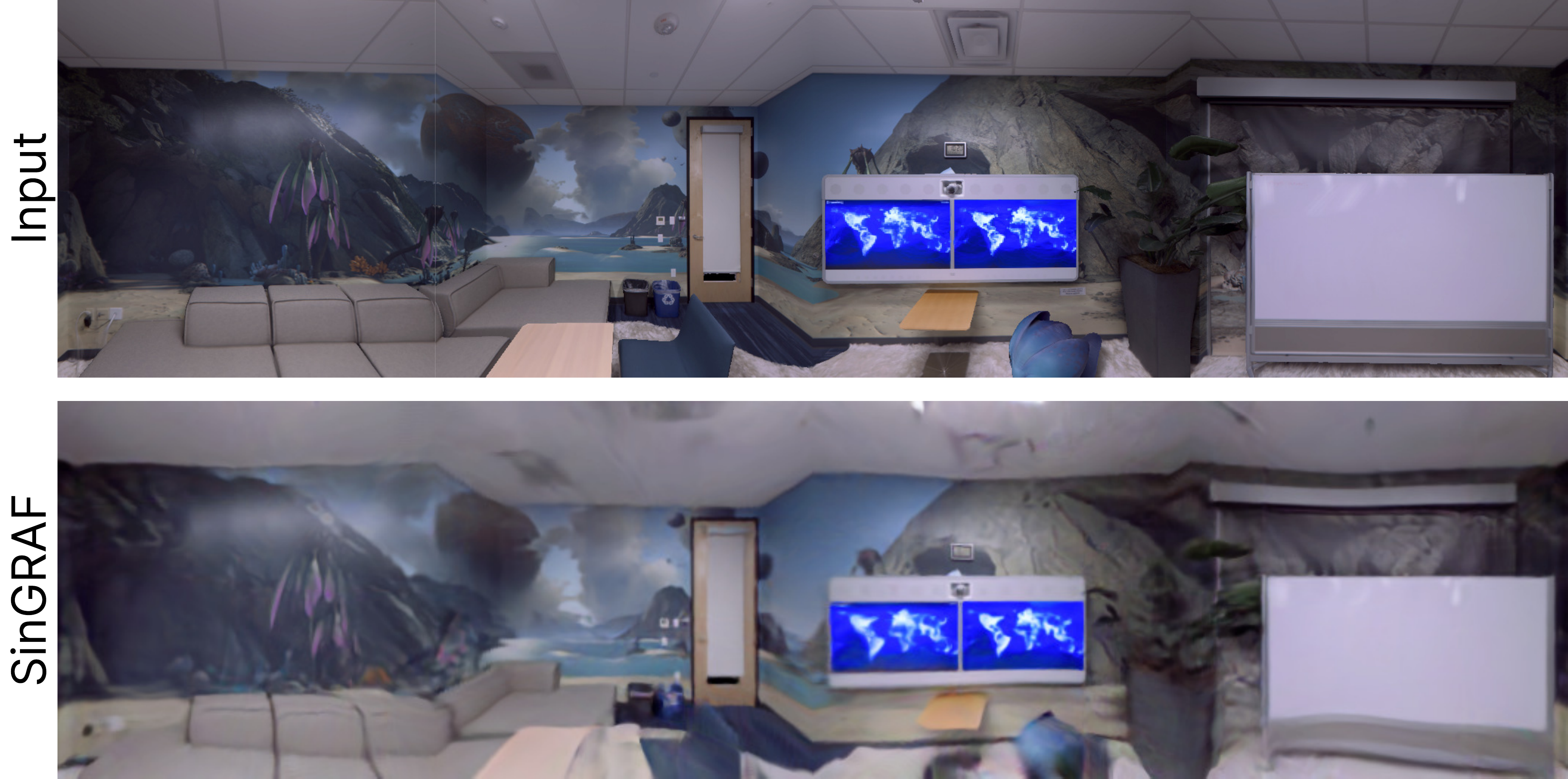}\\
    \caption{Failure case of the ``office\_0'' scene. The lack of diversity in this scene is likely due to the paintings on the walls that uniquely determine the relative locations of most patches. Still, SinGRAF generates high-quality scene which resemble to input.}
    \label{fig:failure_compare}
\end{figure}

We observe that when the scene contains too many local details to be identified from small field-of-view patches, SinGRAF often learns a mode-collapsed latent space. Such an example can be found in Fig.~\ref{fig:failure_compare}, where the scene contains detailed paintings on the wall that uniquely determine the locations of the patches. We note that, while this mode-collapse behavior is unpredictable and thus is a limitation of our approach, the reconstructed scene closely resembles the ground truth scene of the input images. This is surprising, given that our model is given only {\em unposed} images, suggesting a promising future direction toward reconstructing challenging scenes via adversarial training. 

